\PassOptionsToPackage{table}{xcolor}
\documentclass{article} %
\usepackage{iclr2027_conference,times}

\usepackage{amsmath,amsfonts,bm}

\def\eqref#1{equation~\ref{#1}}

\def\1{\bm{1}}

\DeclareMathAlphabet{\mathsfit}{\encodingdefault}{\sfdefault}{m}{sl}
\SetMathAlphabet{\mathsfit}{bold}{\encodingdefault}{\sfdefault}{bx}{n}

\usepackage{hyperref}
\usepackage{url}
\usepackage{booktabs}
\usepackage{graphicx}
\usepackage{amsmath}
\usepackage[table]{xcolor}
\usepackage{multicol}
\usepackage{multirow}
\usepackage[many]{tcolorbox}
\usepackage{longtable}

\definecolor{win}{RGB}{225,242,230}
\definecolor{bad}{RGB}{250,235,232}
\definecolor{accent}{RGB}{61,111,209}
\definecolor{teal}{RGB}{8,117,94}
\definecolor{purple}{RGB}{116,80,184}
\definecolor{ink}{RGB}{24,33,49}
\definecolor{muted}{RGB}{96,108,127}

\newcommand{\msr}{\textsc{MSR}}

\newcommand{\method}{\textsc{Mine--Sample--Render}}

\addtocontents{toc}{\protect\setcounter{tocdepth}{-1}}

\title{Mine, Sample, Render: Test-Time Diversity Alignment of Generative Models}

\author{\textbf{Kostas Triaridis \textsuperscript{1}}\thanks{Work done during a research internship at Microsoft.} \quad
 \textbf{Morris E. Sharp \textsuperscript{2}} \quad
 \textbf{Nebojsa Jojic \textsuperscript{2}}
\\
 \textsuperscript{1}Stony Brook University\quad
 \textsuperscript{2}Microsoft
\\
}

\iclrfinalcopy %
\begin{document}

\maketitle

\begin{abstract}

Modern generative models produce high-quality individual outputs but struggle to sample diversely, even when explicitly instructed to do so. 
We introduce \method \ (MSR), a test-time approach that externalizes the diversity into an explicit, corpus-derived sampling procedure. Our pipeline (i) mines an interpretable categorical schema 
from a corpus of real artifacts and supplies statistics for sampling; (ii) samples attribute configurations from the corpus-derived statistics; (iii) uses a generative model purely 
as a conditional renderer of a sampled configuration. Across weather report generation, user simulation, webpage design, text-to-image generation, and agent-generated 3D assets, we evaluate the diversity and output quality of MSR-guided generation and show that MSR improves alignment with corpus distributions and diversity of generated outputs.
\end{abstract}

\section{Introduction}

Many applications of generative AI require diverse samples, not just individually plausible outputs. Synthetic datasets \citep{huang2025datagen} and simulated user populations should reflect the variation in the populations they represent. Diverse exploration can also improve reasoning and generalization in LLMs \citep{liu2025scaling}, while in text-to-image models, diversity in generations directly drives their test-time scaling capability \citep{rawal2026flash}. Yet modern generators (and even tool-using agents) often sample a much narrower range of variation than what is typically found in real data. This gap has been observed in user simulation \citep{mehri2026measuring}, engineering design \citep{ma2025large}, creative ideation \citep{anderson2024homogenization}, and story generation \citep{xu2025echoes}, and is exacerbated by reward-maximizing post-training \citep[e.g.,][]{shao2024deepseekmath,gao2023scaling} which is commonly used to improve model quality, but can further concentrate samples around high-reward modes and suppress less common but valid alternatives \citep{hu2024amortizing,zhu2026flowrl,liu2026gflowrl}.

\begin{figure}[t]
    \centering
    \includegraphics[width=\linewidth]{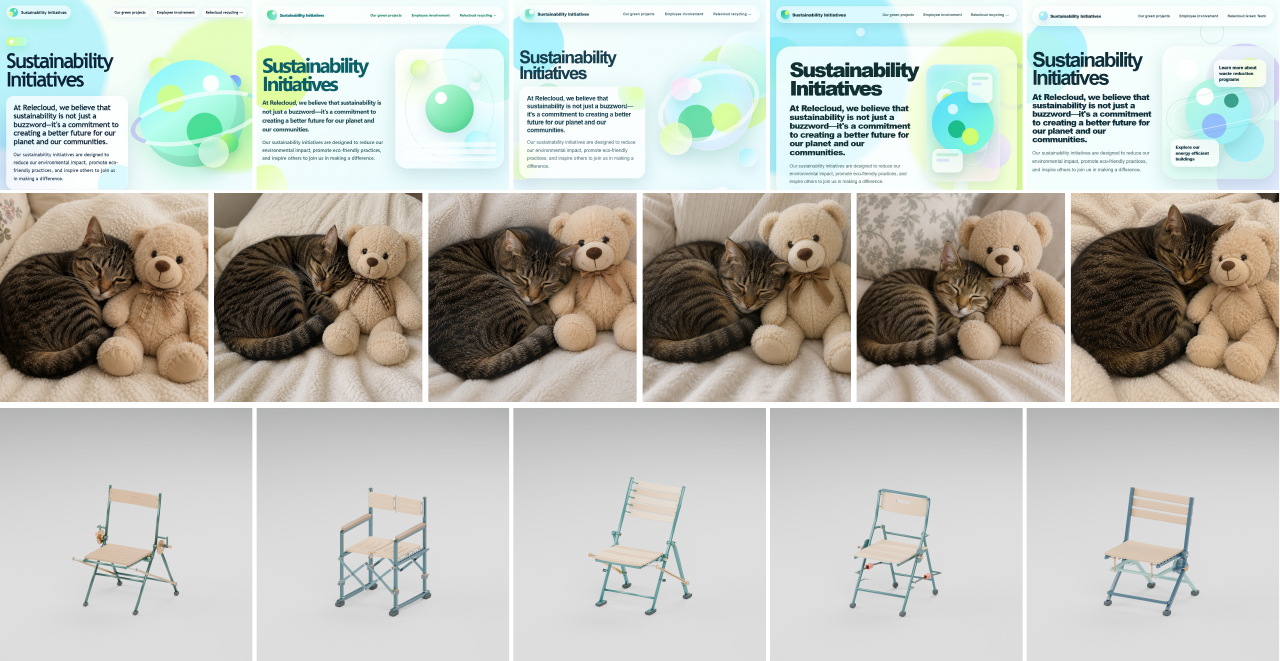}
    \caption{\textbf{Different generations, familiar choices.} Five baseline outputs per task
    . \textbf{Top:} GPT-5.5 webpages generated from the same simple prompt and a given website content. \textbf{Middle:} GPT Image 2 outputs for \textit{``A cat is curled up next to a light teddy bear.''} \textbf{Bottom:} Independent GPT-6 Astra agent runs asked to build a functional 3D folding chair. See Figures~\ref{fig:webpages}, \ref{fig:t2i}, and \ref{fig:chair} for comparison with MSR, with protocols and results in Sections~\ref{webpages}, \ref{t2i}, and \ref{chairs}.}
    \label{fig:intro-homogeneity}
\end{figure}

Changing the prompt {\bf does} change the output, creating an impression of variety, but every prompt leaves choices unspecified, and models repeatedly resolve them in the same way (Figure~\ref{fig:intro-homogeneity}). Simply increasing temperature or asking a model to ``be diverse'' does not identify what should vary, how often, or which combinations are plausible. Even visibly different generations may cover little of the variation in real data, and convincing outputs can therefore be {\bf collectively homogeneous}; even when each user asks for only one. ``Just give me one good webpage'' does not mean ``make it look like everyone else's.'' A single coherent, correct output can still feel obviously AI-generated.

\begin{figure}[t]
  \centering
  \includegraphics[width=\textwidth]{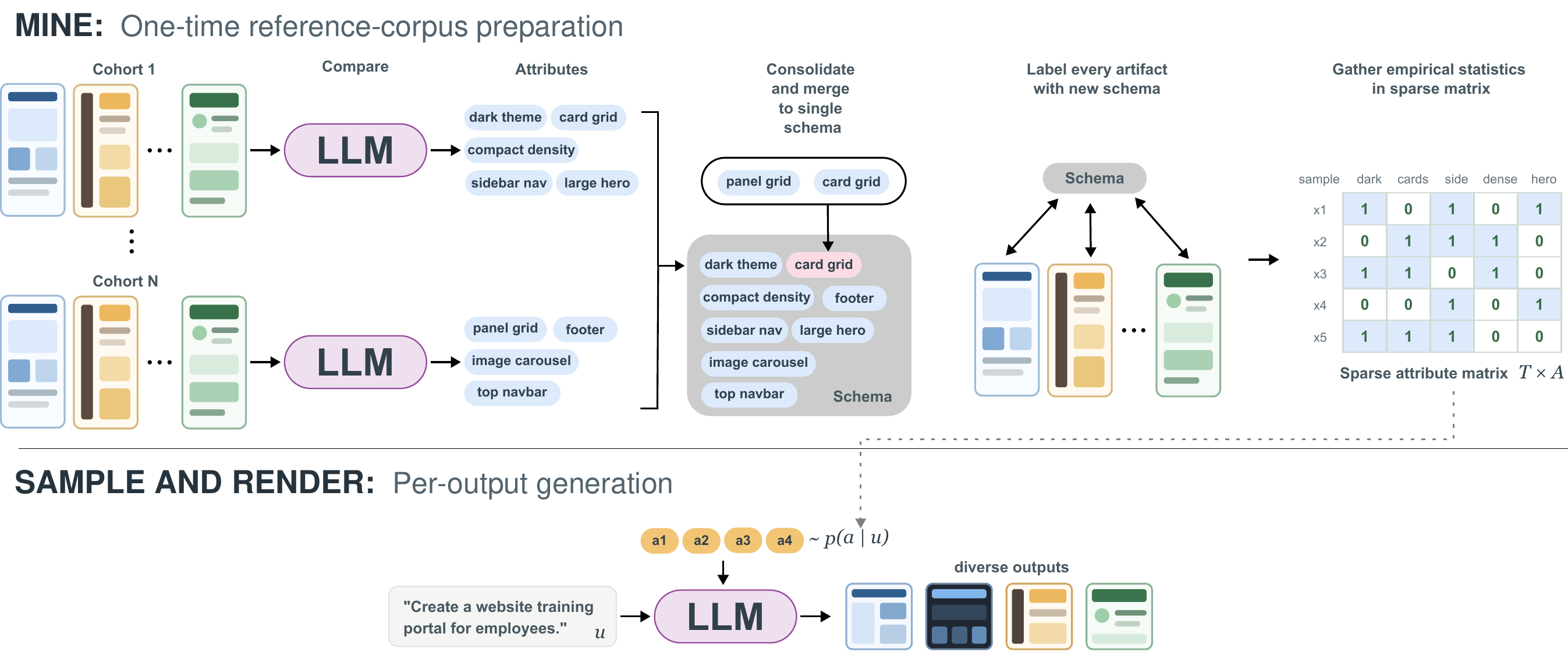}
  \caption{Overview of Mine--Sample--Render. \textbf{(Top)} We use LLMs to mine attributes to create a schema $\mathcal A(\mathcal D)$ by \emph{comparing} small cohorts of artifacts. We then label our source data with the schema to create an empirical distribution of attributes. \textbf{(Bottom)} At inference time, we sample attribute configurations and a model uses them, together with the prompt, to generate diverse outputs.}
  \label{fig:msr-overview}
\end{figure}

We pursue \emph{test-time diversity alignment}: bringing the distribution of generated outputs into closer agreement with the real-data distribution relevant to the task. A reference corpus supplies an empirical target---what varies, how often, and in which combinations. The relevant examples or regularities may already have appeared during pretraining, but that variation is not reliably recovered in the deployed model's samples. Crucially, off-the-shelf models can often identify meaningful differences among real examples and produce the corresponding attributes when explicitly instructed. We introduce \method{} (\msr), shown in Figure~\ref{fig:msr-overview}. In a one-time preprocessing stage for a given reference corpus, \msr{} compares small cohorts of artifacts to discover attributes, then labels the corpus to estimate their joint distribution. To generate a new artifact, it samples an attribute configuration from this distribution and asks a generative model or tool-using agent to render it. This separates \emph{what can be generated} from \emph{how often it should be generated}.

\msr{} relates to attribute-guided generation \citep{yu2023attrprompt,huang2025datagen}, persona conditioning \citep{moon2024virtual,ge2024personahub}, and quality-diversity search \citep{bradley2024qdaif}. Its mining step connects to work on data-induced semantic features and concepts \citep{genz2026,zhong2023d5,lam2024lloom,pham2024topicgpt,wan2024tnt}, while separating structure from realization has precedents in hierarchical text and layout generation \citep{fan2018hierarchical,inoue2023layoutdm}. A close generative precursor is fLSA \citep{xu2025flsa}, which learns natural-language segment tags through reconstruction-based fitting, and samples tag sequences before rendering text. \msr{}'s distinguishing contribution is to discover compositional, artifact-level attributes through direct cohort comparisons and sample their corpus-estimated joint distribution to guide generation. An expanded discussion of related research appears in Appendix~\ref{sec:related-work}.

We apply \msr{} to multiple diverse domains: (i) in a controlled setting of synthetically generated weather reports with known axes of variation, we show that the mining stage can \emph{recover the real axes of variation} that were used in generating the dataset and that the sampling can reproduce their joint distribution faithfully, (ii) in \textbf{synthetic user simulation}, we show that \msr{} reduces the distributional gap between LLM-simulated and real user behavior, (iii) in open-ended \textbf{webpage design with LLMs}, \msr{} greatly improves the diversity of generated webpages over prompt-only baselines, (iv) in \textbf{text-to-image generation}, \msr{} improves the sample diversity of state-of-the-art text-to-image generative models, and (v) in \textbf{3D asset generation} with tool-using agents we show that \msr{} increases both the \emph{diversity} and \emph{quality} of generated assets.

\section{Mine-Sample-Render}
\label{sec:method}

Let \(\mathcal D=\{x_t\}_{t=1}^{T}\) be a corpus of source artifacts that vary along dimensions that are initially unknown. We introduce a mining procedure that produces an interpretable schema $\mathcal A(\mathcal D)= \{a_1,\ldots,a_A\}$ from $\mathcal D$. Attributes are textual descriptions of artifact properties; their presence in source artifacts is recorded by binary labels. Their content and level of abstraction depend on both the mining instructions and what the LLM identifies in the reference corpus. To use images as an example, a schema may therefore mix concrete descriptions, such as ``an English setter'' or ``a child holding an ice cream'', with more general properties, such as ``a low camera angle'' or ``X held by Y'', where X and Y are \emph{role placeholders} for participating entities. At generation time, a sampler \(\phi\) selects an attribute configuration for a prompt \(u\). A rendering procedure $\mathcal Q$ interprets this configuration together with the complete request to produce an artifact:
\[
  \mathbf a\sim \phi(\mathbf a\mid u),
  \qquad
  x\sim \mathcal Q (x\mid u,\mathbf a).
\]

\subsection{Mining attributes by comparison}
\label{sec:mining}

To create the schema $\mathcal A(\mathcal D)$ we use an LLM to mine attributes by \emph{comparing} small cohorts of artifacts. The LLM is given a set of artifacts and asked to identify attributes that are either shared by some artifacts or vary within the cohort.
This process is performed in parallel for a large number of cohorts, resulting in small schemas that were mined independently, possibly containing overlapping or synonymous attributes. We \emph{consolidate} these into a single schema by using an LLM to merge synonymous attributes. We then \emph{back-apply} the consolidated schema to the source corpus, producing a sparse $T\times A$ occurrence matrix from which we estimate attribute frequencies and dependencies.

\subsection{Sampling sparse configurations}
\label{sec:sampling}
We use the occurrence matrix to estimate a joint distribution over attributes and sample configurations with \(\phi\). We explore two approaches: topic-based sampling and sequential ancestral sampling, which conditions each draw on preceding assignments.

\paragraph{Conditioning on a request.}
Users supply ordinary-language requests without needing to know the attribute vocabulary. For a small schema, an LLM could identify required and compatible attributes directly from their descriptions; we do not evaluate that alternative here. For large schemas, we augment the statistical representation with \emph{lexical features}: TF--IDF-selected words and shared n-grams from attribute names and descriptions. Each source artifact inherits features from its present attributes, supplemented by caption terms when available. These shared features provide conditioning evidence when the request does not match a complete attribute description. Both mined attributes and lexical features can condition sampling. Explicit requirements, such as left-hand navigation in a webpage, may be fixed; compatibility identifies eligible candidates, while lexical matches supply statistical evidence. The sampled configuration contains mined attributes only.

\subsubsection{Sampling with a topic model}
\label{topic}
We form a joint association matrix over mined attributes and lexical features. For the $N$ retained features, we calculate Normalized Pointwise Mutual Information (NPMI): $
\operatorname{NPMI}(i,j)
=
\frac{\log \frac{P(i,j)}{P(i)P(j)}}{-\log P(i,j)}
$.
We set negative values to zero to form a positive-NPMI matrix $A^+$ and fit $K$ topics using symmetric nonnegative matrix factorization, such that $A^+ \approx HH^\top$, where $H \in \mathbb R^{N \times K}$. We infer topic probabilities $p(t\mid u)$ from matched conditioning features and normalize each topic's mined-attribute loadings to obtain $p(a\mid t)$. We first draw a topic $t$, then a mined attribute $a$ from that topic, with $p(a\mid u)=\sum_{t=1}^{K}p(a\mid t)p(t\mid u)$ and repeat until the desired number of attributes has been selected.

\subsubsection{Ancestral sampling}
\label{ancestral}
With conditioning on \(u\) implicit, we start with a random ordering \(\pi\), and sample each attribute $a_{\pi_j}$ using its conditional probability given preceding assignments $p(a_{\pi_j}\mid a_{\pi_1},\ldots,a_{\pi_{j-1}})$. In the image implementation, prompt-derived lexical evidence reweights these empirical property probabilities.
As $j$ increases, empirical conditionals have progressively less--and often no--matching support; and the sampler would reproduce only combinations present in the source dataset. In order for our  
sampling to be able to generalize beyond our training data, we remove sampled attributes from the conditioning set until sufficient support remains. We rank the previously sampled attributes by their mutual information with $a_{\pi_j}$, and select the first $N$ so that at least \(m\) source samples remain.

We also experiment with a mixture of ancestral and topic-based sampling, where each attribute is sampled using $
p_\lambda(a\mid u)=(1-\lambda)p_{\text{ancestral}}(a\mid u)+\lambda p_{\text{topic}}(a\mid u)$, where $\lambda$ is a hyperparameter.

\section{Experiments}
\label{sec:experiments}

\subsection{Attribute Recovery from Weather Reports}
We first want to evaluate MSR on a dataset where the axes of variation are controlled and known, and evaluate (i) whether the mining stage can correctly recover the true axes of variation, and (ii) whether sampling with the induced schema can produce a distribution close to the source one. We create a dataset of synthetically generated weather reports based on \textbf{real weather data}: we obtain daily weather conditions for ten cities in 2016, and discretize them into binary attributes related to \textit{temperature}, \textit{sky}, \textit{wind}, and \textit{precipitation}, producing a data-generating schema $\mathcal A_{\text{source}}$. We use the real attribute configurations to prompt GPT-5-mini to generate a dataset of weather reports. We then evaluate the mining stage of MSR: we use GPT-5-mini to mine a schema by comparing sampled reports from our generated dataset, discovering a new schema $\mathcal A$, which is almost identical to the data-generating schema $\mathcal A_{\text{source}}$, showing that MSR can effectively recover the true axes of variation by comparing data samples and identifying their similarities and differences (Appendix Section~\ref{app:weather-mining}).

\begin{figure}[h]
\begin{minipage}{0.45\textwidth}
        \centering
        \includegraphics[width=\linewidth]{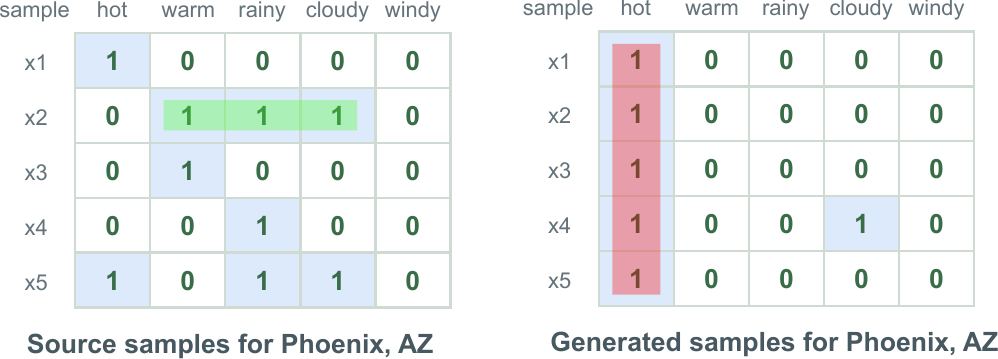}
        \label{fig:toy_msr}
    \end{minipage}
\begin{minipage}{0.5\textwidth}
\small
\label{tab:weather-heldout}
\begin{tabular}{lccc}
\toprule
\textbf{Method}
& \textbf{Fwd KL $\downarrow$}
& \textbf{JS $\downarrow$}
& \textbf{Coverage $\uparrow$} \\
\midrule
Prompt only & 0.570 & 0.484 & 21\% \\
Append reports & 0.394 & 0.342 & 24\% \\
\rowcolor{win}
\textbf{MSR} & \textbf{0.181} & \textbf{0.203} & \textbf{68\%} \\
\midrule
Real distribution & 0.150 & 0.166 & 83\% \\
\bottomrule
\end{tabular}
\end{minipage}
\caption{\textbf{(Left)} LLMs generate weather reports that \colorbox{red}{collapse to a specific mode} and do not produce samples that follow conditions that are \colorbox{green}{common but do not belong to the dominant mode}, even when given the source dataset in the context \textbf{(Right)}. Generated-text distributional gap against the held-out real distribution of weather conditions.}
\end{figure}

We then sample attribute configurations from our induced schema and use them to generate weather reports. We label generated reports using the original weather schema and compare their joint configurations with held-out 2017 observations, reporting Forward KL, Jensen--Shannon divergence, and support coverage. Baselines generate reports either from the prompt alone or with the full source-report corpus in context. A reference condition samples the data-generating schema's joint distribution and renders the resulting configurations, separating schema-estimation error from rendering and distribution-shift effects. Table~\ref{tab:weather-heldout} shows that MSR reduces Forward KL from
0.570 to 0.181 and JS from 0.484 to 0.203, while increasing coverage from
21\% to 68\%.

\subsection{User simulation}

We apply \method{} to the task of using LLMs to simulate users that interact with LLM assistants. Given a high-level ``user goal'', an LLM generates a plausible user trajectory, while another LLM provides assistant responses. In the mining stage, we compare real user-LLM interactions and mine attributes that describe user behavior. Then we sample an attribute configuration and use it to condition a \emph{user simulator}, which interacts with an \emph{assistant}. We evaluate the gap between a held-out distribution of real users and a simulated distribution of users with the same ``user goal'' using the evaluation protocol introduced by \citet{mehri2026measuring}.

\paragraph{Implementation.} We focus on simulating users for coding tasks and use the same dataset of 5000 real coding conversations used by \citet{mehri2026measuring} for evaluation. We mine a distinct set of 600 real coding conversations from WildChat \citep{zhao2024wildchat} to use for schema generation. 
We use 2 different mining prompts to recover attributes at different granularities, resulting in 2 schemas with 84 (compact) and 465 (fine-grained) attributes, respectively. For our experiments, we use GPT-5-mini as the miner. We fix GPT-5.4 as the assistant and compare four user-simulation conditions: prompt only, using a coding-relevant PersonaHub persona \citep{ge2024personahub}, and sampled attribute configurations with both schemas. We specifically filter for personas suited to coding tasks and sample from this subset uniformly with replacement when simulating users. More details on the evaluation implementation are in Appendix~\ref{app:user-simulation}.

\begin{table}[h]
\centering
\caption{Distributional gap between real and simulated user behaviors in coding, following the evaluation protocol of \citet{mehri2026measuring}: an annotator describes each real and simulated conversation along behavioral facets. These descriptions are embedded and clustered, and the resulting real and simulated cluster distributions are compared using Forward and Backward KL and Jensen--Shannon divergence. Coverage is calculated as the percentage of real clusters with at least one simulated user.}

\label{tab:user_sim}
\begin{tabular}{l r r r r}
\toprule
\textbf{User simulator}
& \textbf{$\mathrm{KL}_{\mathrm{fwd}}\downarrow$}
& \textbf{$\mathrm{KL}_{\mathrm{bwd}}\downarrow$}
& \textbf{JS $\downarrow$}
& \textbf{Coverage $\uparrow$} \\
\midrule

\textbf{GPT-5-mini} (prompt only)
& 0.5668 & 0.5764 & 0.5866 & 0.4646 \\
\quad $+$ PersonaHub (coding-filtered)
& 0.5971 & 0.6033 & 0.6157 & 0.4013 \\
\rowcolor{win}
\quad $+$ \textsc{MSR} (compact)
& 0.5236 & 0.5556 & \textbf{0.5319} & \textbf{0.5649} \\
\rowcolor{win}
\quad $+$ \textsc{MSR} (fine-grained)
& \textbf{0.5184} & \textbf{0.5339} & 0.5373 & 0.5106 \\
\midrule

\textbf{GPT-5.5} (prompt only)
& 0.5258 & 0.5272 & 0.5167 & 0.6345 \\
\quad $+$ PersonaHub (coding-filtered)
& 0.5395 & 0.5505 & 0.5371 & 0.5850 \\
\rowcolor{win}
\quad $+$ \textsc{MSR} (compact)
& \textbf{0.3101} & \textbf{0.3272} & \textbf{0.3086} & \textbf{0.7813} \\
\rowcolor{win}
\quad $+$ \textsc{MSR} (fine-grained)
& 0.3134 & 0.3324 & 0.3157 & 0.7597 \\
\midrule

\textbf{GPT-4o} (prompt only)
& 0.6049 & 0.6101 & 0.6362 & 0.2987 \\
\quad $+$ PersonaHub (coding-filtered)
& 0.6188 & 0.6212 & 0.6530 & 0.2557 \\
\rowcolor{win}
\quad $+$ \textsc{MSR} (compact)
& \textbf{0.3530} & 0.3887 & \textbf{0.3664} & 0.7298 \\
\rowcolor{win}
\quad $+$ \textsc{MSR} (fine-grained)
& 0.3565 & \textbf{0.3848} & 0.3731 & \textbf{0.7324} \\

\midrule

\textbf{Qwen 3.8 27B} (prompt only)
& 0.5914 & 0.5991 & 0.6071 & 0.4018\\
\quad $+$ PersonaHub (coding-filtered)
& 0.5971 & 0.6023 & 0.6168 & 0.3571 \\\rowcolor{win}
\quad $+$ \textsc{MSR} (compact)
& 0.3306 & 0.3515 & 0.3284 & 0.7817 \\
\rowcolor{win}
\quad $+$ \textsc{MSR} (fine-grained)
& \textbf{0.3062} & \textbf{0.3354} & \textbf{0.3125} & \textbf{0.8122} \\
\midrule

\textbf{Kimi K2.6} (prompt only)
& 0.5895 & 0.6105 & 0.6086 & 0.3927 \\
\quad $+$ PersonaHub (coding-filtered)
& 0.5914 & 0.6232 & 0.6057 & 0.4217 \\
\rowcolor{win}
\quad $+$ \textsc{MSR} (compact)
& \textbf{0.2962} & \textbf{0.3248} & \textbf{0.2975} & 0.7783 \\
\rowcolor{win}
\quad $+$ \textsc{MSR} (fine-grained)
& 0.3008 & 0.3265 & 0.3153 & \textbf{0.7798}\\

\bottomrule
\end{tabular}
\end{table}

\paragraph{Results.} We evaluate a range of proprietary and open-weight LLMs as user
simulators: GPT-5 mini \citep{openai2025gpt5},
GPT-5.5 \citep{openai2026gpt55},
GPT-4o \citep{openai2024gpt4o},
Qwen3.8-27B \citep{qwen2026qwen38},
and Kimi K2.6 \citep{moonshot2026kimik26}. As shown in Table~\ref{tab:user_sim}, MSR greatly improves all user simulators, while sampling personas from PersonaHub fails to capture meaningful variation in user behavior, as simulators collapse to similar behavioral patterns with limited diversity.

\subsection{Webpage Generation}
\label{webpages}

\begin{figure}[t]
    \centering
    \includegraphics[width=\linewidth]{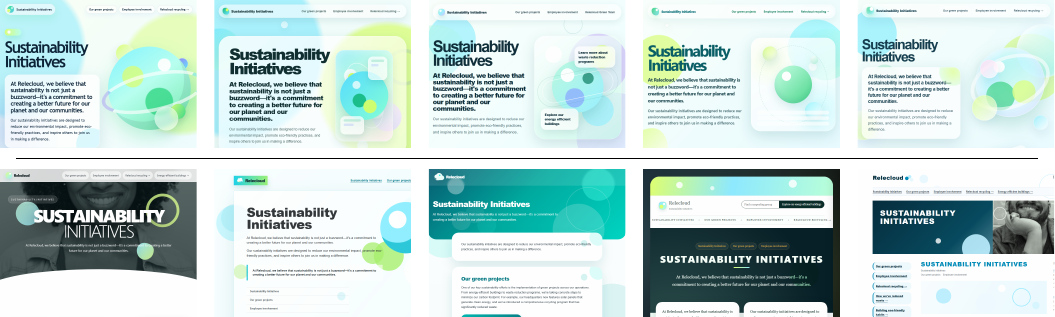}
    \caption{Webpages generated by GPT-5.5 with a fixed prompt (\emph{top}) and additionally MSR (\emph{bottom}).}
    \label{fig:webpages}
\end{figure}

Prior work reports convergence in the visual appearance of user interfaces produced by LLM-based generation tools \citep{imteyaz2026designtheater}, while practitioner accounts describe recurring aesthetic and design defaults in LLM-generated webpages \citep{anthropic2025frontend}. We make the same observation: LLM-generated webpages tend to collapse to a very specific look and design structure (Figure~\ref{fig:webpages}). In this setting, we apply \method{} to improve the diversity of generated pages given some fixed provided content. In the mining stage, a VLM is given the HTML code and a screenshot for a cohort of webpages and asked to compare them to produce a schema of design attributes. For sampling, we sample attributes with ancestral sampling (Section~\ref{ancestral}) while also identifying and including attributes specified by the initial prompt using an LLM.

\paragraph{Implementation.} We use two different datasets to mine a schema to assess how data complexity and scale affect its quality and usefulness. We mine the first schema of 485 attributes using GPT-5-mini from 10,000 webpages sourced from WebSight v0.1 \citep{laurençon2024websight}, which contains simple, synthetically generated webpages. For the second schema, we use the 484 more realistic and complex webpages in the Design2Code dataset \citep{si2025design2code} and mine 5,484 attributes using GPT-5.6 Sol. For evaluation, we use 100 webpage authoring prompts paired with grounding documents that contain the desired content (images, text), and GPT-5.5 or Kimi K2.6 as the generator. As a baseline, we prompt an LLM generator with just the website authoring prompt and the grounding documents. For each prompt, we generate 10 webpages with each method and measure the diversity of the generated sets of webpages using the Vendi Score \citep{friedman2023vendi} in two different settings: (i) we render screenshots of the generated webpages, embed them using DINOv2 \citep{oquab2024dinov2}, and calculate the Vendi Score based on the similarity of the image embeddings; (ii) we strip the HTML pages of their content, keeping only the structural elements as text, embed them using OpenAI's text-embedding-3-large model, and calculate the Vendi Score based on the similarity of the text embeddings. More implementation details are included in the Appendix Section~\ref{app:webpage-generation}. 

\begin{table}[h]
\centering
\small
\caption{Diversity evaluation for generated webpages. 
}

\label{tab:webpage_vendi}
\begin{tabular}{lcccc}
\toprule
\multirow{2.5}{*}{\textbf{Webpage generator}} &\multicolumn{2}{c}{\textbf{Vendi Score}}&\multicolumn{2}{c}{\textbf{Odd-one-out}}\\
\cmidrule(lr){2-3}
\cmidrule(lr){4-5}

& text$\uparrow$
& image$\uparrow$
& \% removed$\uparrow$
& mean removal rank$\downarrow$ \\
\midrule

\textbf{GPT-5.5} (prompt only)
& 2.158 & 2.446 & - & - \\
\rowcolor{win}
\quad $+$ \textsc{MSR} (WebSight)
& \textbf{3.956} & \textbf{4.116} & 95.075\% & 6.20 vs 19.30 \\
\rowcolor{win}
\quad $+$ \textsc{MSR} (Design2Code)
& 3.273 & 3.457 & 97.650\% & 5.77 vs 19.73 \\
\midrule

\textbf{Kimi K2.6} (prompt only)
& 2.426 & 3.150 & - & -\\
\rowcolor{win}
\quad $+$ \textsc{MSR} (WebSight)
& \textbf{3.842} & \textbf{4.077} & 75.575\% & 9.16 vs 16.33\\
\rowcolor{win}
\quad $+$ \textsc{MSR} (Design2Code)
& 3.267 & 3.235 & 80.675\% & 8.36 vs 17.13\\

\bottomrule
\end{tabular}
\end{table}

\paragraph{Odd-one-out} We also introduce an odd-one-out style evaluation \citep{chang2009reading} using a VLM judge. For each prompt and generator, we pool ten prompt-only and ten MSR-generated screenshots of webpages under neutral labels. The judge selects the webpage whose visual design is most unlike the other pages in the pool. We remove the selected page, reshuffle the pages, and repeat until only 10 pages remain. We run four repetitions of the process with each one shuffled independently. Removed pages receive ranks 1--10, while all survivors receive
rank 20. We report the mean removal rank for each method as well as the share of all removals (\% removed). Lower ranks and higher removal shares indicate more distinct samples, and as such greater diversity within the mixed pool.

\paragraph{Results.} Table~\ref{tab:webpage_vendi} shows that schema-conditioned pages increase diversity (both visual and structural) compared to a prompt-only baseline for all models and both schemas. Schema-conditioned pages showcase higher Vendi Scores and account for the majority of odd-one-out removals in every comparison: 95.1--97.7\% for GPT-5.5 and 75.6--80.7\% for Kimi K2.6. The MSR-generated pages are more stylistically diverese, as shown in Figure~\ref{fig:webpages}. For Kimi K2.6, Vendi Score and Odd-one-out both point towards an increase in diversity over baseline using MSR, but a smaller gain compared to GPT 5.5. This is potentially due to differences in the model capabilities. For example, Kimi K2.6 does not consistently follow our instruction to avoid JavaScript on webpages, and we had to add additional emphasis in the page-generation prompts to prevent it from ignoring instructions.

\subsection{Text-to-Image Generation}
\label{t2i}

\begin{figure}[t]
    \centering
    \includegraphics[width=\linewidth]{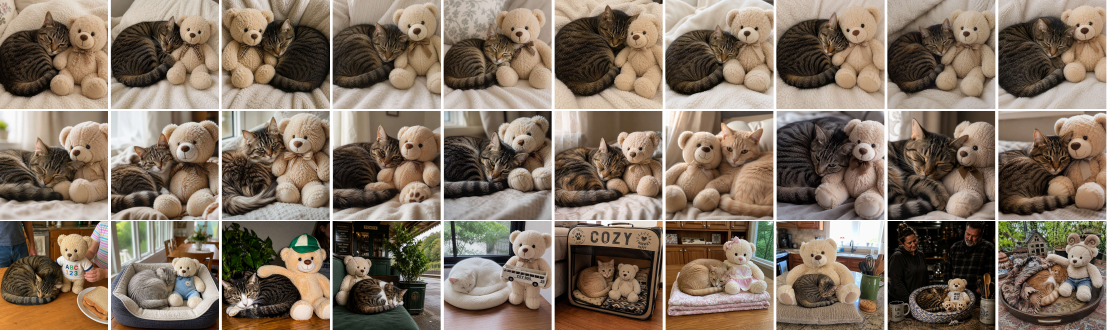}
    \caption{Images generated by GPT Image 2 using the prompt \textit{``A cat is curled up next to a light teddy bear.''} \textbf{(Row 1)} Original prompt only. \textbf{(Row 2)} Independent rewrite with GPT-5.6 Sol. \textbf{(Row 3)} \textsc{MSR}--Resolved, using GPT-5.6 Sol for semantic resolution.}
    \label{fig:t2i}
\end{figure}

We test whether MSR increases image diversity for fixed text requests beyond what can be obtained through prompt rewriting, and whether the gains transfer across image generators (Figure~\ref{fig:t2i}).

\paragraph{Implementation.} We use up to 100 images from each category of the training set of MS-COCO \citep{lin2014coco} and mine a schema of 28,875 properties using GPT-5.6 Sol. We fit a 20-topic model on a compact, support-filtered vocabulary of 350 mined attributes and 450 lexical features. The topic component captures shared modes of variation rather than attempting to preserve the full range of rare attributes and their specific dependencies. The ancestral component retains access to the full schema through a request-specific pool of at most 250 attributes, allowing more specific details to be sampled when supported by the corpus. We sample with a mix between topic and ancestral sampling with $\lambda=0.5$. Topic draws introduce broad variation, while ancestral draws condition on previously selected attributes---including those drawn from topics---to introduce empirically compatible details. 
Feature-selection criteria and sampling details are provided in Appendix~\ref{app:t2i}.

We evaluate GPT Image 2, MAI-Image-2.5, and FLUX.2-flex on 50 held-out MS-COCO validation captions, one from each of 50 categories, with ten images per caption and arm. We compare five arms: prompt only, independent rewrite, coordinated rewrite, \textsc{MSR}--Direct, and \textsc{MSR}--Resolved. Prompt rewriting is widely used in text-to-image generation \citep{betker2023improving, esser2024scaling, deng2025emerging} and evaluation \citep{kamath2025geneval}. Our independent baseline rewrites the original request in separate LLM calls; the coordinated baseline produces a set of rewrites in one call aiming to improve their diversity. \textsc{MSR}--Direct provides the sampled attributes to the generators directly, while \textsc{MSR}--Resolved passes them to an LLM that first resolves placeholders and composes coherent scene instructions. Both variants may omit attributes that cannot be realized plausibly alongside the request and other retained attributes. GPT-5.6 Sol performs both baseline rewriting and MSR resolution. We reuse the prompts, sampled configurations, and resolved instructions across generators, apart from one documented provider-specific safety replacement (Appendix~\ref{app:t2i}).

\paragraph{Evaluation.} We compute the Vendi Score \citep{friedman2023vendi} separately for each ten-image set using CLIP ViT-L/14 \citep{radford2021learning} and DINOv2 ViT-G/14 \citep{oquab2024dinov2} embeddings with fixed RBF kernels, then average over the 50 requests. We also conduct a blinded odd-one-out evaluation. For each request and generator, we pool ten images from each of the three baselines and \textsc{MSR}--Resolved. GPT-5.6 Sol selects the image most unlike the others; we remove it and reshuffle the remaining pool, repeating for 30 rounds and leaving ten survivors. We report the fraction removed and mean removal rank, assigning rank 40 to survivors. \textsc{MSR}--Direct is not included in this four-arm evaluation. We also evaluate prompt satisfaction, visual acceptability and attribute realization in Appendix Section \ref{app:t2i}.

\paragraph{Results.} Both MSR variants produce more diverse sets than prompt-only generation, independent rewriting, and coordinated rewriting across all three generators (Table~\ref{tab:t2i_vendi}). Semantic resolution further increases mean diversity: DINOv2 Vendi scores reach 6.68--6.77, compared with 4.32--4.55 for coordinated rewriting. The blinded odd-one-out evaluation agrees: \textsc{MSR}--Resolved has the lowest mean removal rank on all 50 requests for each generator. All 500 resolved images are removed for GPT Image 2 and FLUX.2-flex, and 499 of 500 for MAI-Image-2.5.

\begin{table}[h]
\centering
\caption{Image diversity across three generators. Vendi scores are averaged over 50 ten-image sets per arm. Odd-one-out compares the three baselines with \textsc{MSR}--Resolved, removing 30 of 40 images per request; survivors receive rank 40. Dashes indicate that Direct was not included in this evaluation.}

\label{tab:t2i_vendi}
\begin{tabular}{lcccc}
\toprule
\multirow{2.5}{*}{\textbf{Generator / conditioning}} &\multicolumn{2}{c}{\textbf{Vendi Score}}&\multicolumn{2}{c}{\textbf{Odd-one-out}}\\
\cmidrule(lr){2-3}
\cmidrule(lr){4-5}

& CLIP$\uparrow$
& DINOv2$\uparrow$
& \% removed$\uparrow$
& mean removal rank$\downarrow$ \\
\midrule

\textbf{GPT Image 2} (prompt only)
& 2.37 & 2.16 & 13.80\% & 37.97\\
\quad $+$ {independent rewrite}
& 3.02 & 3.19 & 88.80\% & 25.26\\
\quad $+$ {coordinated rewrite}
& 3.86 & 4.32 & 97.40\% & 16.45\\
\rowcolor{win}
\quad $+$ \textsc{MSR}--Direct
& 5.14 & 4.97 & - & -\\
\rowcolor{win}
\quad $+$ \textsc{MSR}--Resolved
& \textbf{6.68} & \textbf{6.73} & \textbf{100.00\%} & \textbf{6.82}\\
\midrule

\textbf{MAI-Image-2.5} (prompt only)
& 3.23 & 3.77 & 61.8\% & 28.86 \\
\quad $+$ {independent rewrite}
& 3.01 & 3.44 & 51.40\% & 31.35 \\
\quad $+$ {coordinated rewrite}
& 3.83 & 4.55 & 87.00\% & 19.59 \\
\rowcolor{win}
\quad $+$ \textsc{MSR}--Direct
& 5.55 & 5.61 & - & - \\
\rowcolor{win}
\quad $+$ \textsc{MSR}--Resolved
& \textbf{6.37} & \textbf{6.68} & \textbf{99.80\%}& \textbf{6.70}\\
\midrule

\textbf{FLUX.2-flex} (prompt only)
& 2.64 & 3.25 & 54.00\% & 30.84 \\
\quad $+$ {independent rewrite}
& 2.75 & 3.44 & 58.00\% & 30.01 \\
\quad $+$ {coordinated rewrite}
& 3.30 & 4.44 & 88.00\% & 19.19 \\
\rowcolor{win}
\quad $+$ \textsc{MSR}--Direct
& 6.17 & 6.29 & - & - \\
\rowcolor{win}
\quad $+$ \textsc{MSR}--Resolved
& \textbf{6.36} & \textbf{6.77} & \textbf{100.00\%} & \textbf{6.46} \\

\bottomrule
\end{tabular}
\end{table}

\subsection{Agentic 3D asset Generation}
\label{chairs}

Previous experiments were focused on improving the diversity of outputs of individual generative models such as LLMs or text-to-image generative models. Here we ask a more demanding question: when a capable model can plan, write code, use tools, and iterate and revise generated outputs, does it nevertheless repeatedly choose a narrow range of designs? To evaluate this, we introduce a task where an agent is asked to create a fully functional folding chair in 3D, including all necessary mechanical components, and to demonstrate its functionality. We choose the folding chair because it is a complex design assignment with a very diverse range of possible design choices (materials, colors, texture, proportions, detailing, etc.), but those choices must result in a valid structure and a meaningful folding transformation (armrests affect clearance, pivots and braces must remain connected as the chair closes, etc.). We use GPT-6 Astra as a tool-using design agent and evaluate the validity and diversity of the proposed designs. We gather a set of 160 folding chair designs, with images from multiple sources (more details in Appendix Section \ref{3d:data}) and use GPT 5.6 Sol to mine a schema of 447 attributes. We sample attribute sets, using ancestral sampling, as described in Section~\ref{ancestral}. After sampling an attribute set, an agent is instructed to create a functional folding chair in 3D, without importing a chair or component models; to save a reproducible source, an editable Blender scene, a GLB, and demonstration outputs; and to record any references used. We generate a set of 20 chairs and compare MSR with two baselines: (i) an agent independently creating a chair 20 times, and (ii) an agent is asked to create 20 chairs, each different than the previous one, retaining the context. For evaluation, we use the odd-one-out method described in Section~\ref{webpages}, using GPT-6 Astra and rendered images of the generated chairs with a fixed camera, lighting, and background. We then design an evaluation rubric, where an agent is asked to rate each generated chair on its validity and quality, using the generated images, detail views, and folding demonstrations. More details about the evaluation are included in the Appendix Section~\ref{3d:eval}.

\begin{figure}[t!]
    \centering
    \includegraphics[width=\linewidth]{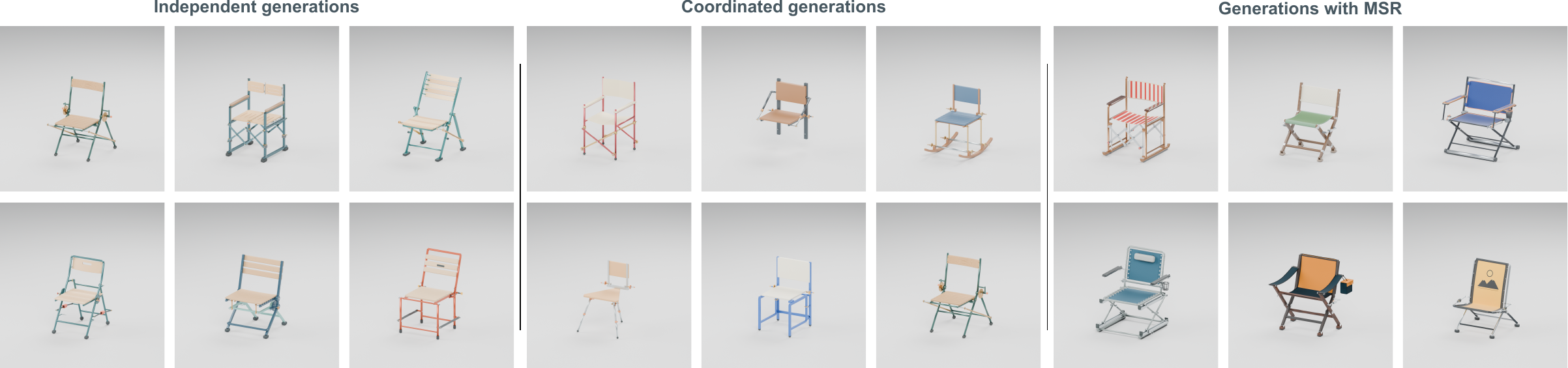}
    \caption{3D folding chairs generated with GPT-6 Astra.}
    \label{fig:chair}
\end{figure}

\begin{table}[h]
\centering
\caption{Evaluation for 3D folding chair generation.}

\label{tab:chairs}
\begin{tabular}{lcc}
\toprule
\textbf{Method}
& \textbf{Mean odd-one-out removal rank $\downarrow$}
& \textbf{Quality $\uparrow$} \\
\midrule

GPT-6 Astra independent generations
& 48.800 & 82.35 \\
GPT-6 Astra coordinated generations
& 25.125 & 70.80 \\
\rowcolor{win}
GPT-6 Astra $+$ \textsc{MSR}
& \textbf{17.100} & \textbf{83.55} \\
\bottomrule
\end{tabular}
\end{table}

\paragraph{From attributes to an articulated model.}
Each brief supplies 63--111 attributes describing parts, materials, and construction. Because these attributes omit dimensions and mechanisms, the agent must interpret them to engineer a functional, articulated 3D asset. It authors a Blender scene, designs the mechanical joints, programs the complete folding sequence, and checks physical clearances across intermediate poses (Figure~\ref{fig:chair_folding}). If sampled properties mechanically conflict, the agent modifies the design and logs the changes. The agent is instructed to produce reproducible source code, an editable Blender scene, an animated GLB, and rendered demonstrations. An example of this output is provided in Appendix Section~\ref{3d:output}.

\paragraph{Results.} In Table~\ref{tab:chairs} we demonstrate that GPT-6 can generate high-quality 3D folding chairs, but shows limited variation in designs (Figure~\ref{fig:chair}). Instructing the agent to make generations that are dissimilar to previous ones leads to some increase in the variety of outputs but a degradation in generation quality. MSR overcomes this limitation by systematically navigating the design space. Our results show that GPT-6 using MSR successfully generates the most diverse set of outputs without any decrease in generation quality.

\section{Conclusion}
While modern generative models excel at producing high-quality individual outputs, they inherently struggle to generate diverse outputs and reproduce the distribution of real populations. To address this limitation, we introduced \method\ (MSR), a training-free framework that decouples the statistical sampling of implicit generative choices from the generation step itself. By mining interpretable categorical schemas from real-world corpora and explicitly sampling attribute configurations, MSR allows the underlying model to act purely as a conditional renderer. Our evaluations across highly varied domains—spanning weather reports, user simulations, web design, image generation, and agent-created 3D assets—demonstrate that MSR significantly enhances output diversity and aligns generated populations with real distributions. Ultimately, MSR provides a robust, interpretable mechanism for ensuring generative models can capture the full richness and variability of real-world data.

\subsection*{AI use statement}

In this work, we used generative AI tools to draft sections of the paper, to find related work, and to aid or polish writing. Specifically, we created initial drafts of implementation detail sections by providing an LLM with the source code and requesting a text report. Authors verified the provided text and edited it for clarity. Additionally, we used LLMs to find and gather related literature, which the authors reviewed, and to rewrite parts of the paper to improve clarity. We also used generative AI tools for generating synthetic data sets and implementing methods. We implemented our method using coding agents and reviewed and verified the generated code for correctness. The dataset of 100 prompts and grounding documents used for evaluating webpage generation was generated with an LLM and filtered by the authors.

We have not used generative AI tools to help develop theoretical models or conceptual frameworks, formulate mathematical claims, provide critical ingredients for proving mathematical claims, assist in the writing of proofs, propose or refine hypotheses, design or provide feedback on research methodology or experiments, assist with translation, support qualitative and thematic data analysis, or interpret results. 

We have reviewed all AI-assisted work. We take responsibility for the final content of this work, including text, claims or artifacts produced with the aid of generative AI.

\bibliography{iclr2027_conference}
\bibliographystyle{iclr2027_conference}

\newpage
\appendix
\section*{Appendix}
\addtocontents{toc}{\protect\setcounter{tocdepth}{3}}
\tableofcontents

\clearpage

\section{Related Work}
\label{sec:related-work}

\paragraph{From explicit attributes to data-induced semantic variables.}

A common temptation when using foundation models is to ask them to perform both
semantic and statistical reasoning in one unconstrained call: invent the latent
variables, decide which ones matter, determine how they are distributed, and
generate an artifact. AttrPrompt and DataGen instead condition generation on
explicit attributes obtained through user specification, LLM prior knowledge,
or selected in-context examples \citep{yu2023attrprompt,huang2025datagen}, but
do not estimate the source population's joint distribution: diversity comes
from random attribute combinations, LLM prompting, or post-hoc filtering. Persona Hub conditions generation on sampled persona records, while QDAIF iteratively searches a user-specified behavior space for high-quality distinct outputs \citep{ge2024personahub,bradley2024qdaif}; these approaches likewise externalize diversity without estimating a target corpus's joint distribution over compositional attributes. This
matters because the semantic diversity potentially encoded during training is
not reliably exposed at generation time; repeated LLM outputs reproduce a
narrow set of plot patterns compared with human corpora and human ideation or design processes \citep{xu2025echoes,anderson2024homogenization,ma2025large}.
We therefore treat the corpus as external memory: the foundation model
rediscovers human-readable dimensions by comparing observed artifacts, while
conventional statistics estimates how those dimensions are distributed. This
division follows two hybrid statistical--foundational models that mine semantic
features through generalized EM but optimize different likelihoods.

GenZ \citep{genz2026} optimizes the conditional likelihood $p(y\mid x)$,
linking a semantic item $x$ to an external, possibly high-dimensional numerical
target $y$ through latent semantic features. Its statistical model uses target
residuals to identify a useful split; a foundation model names the distinction
and applies that natural-language question to all items. Refitting tests the
resulting feature, and contraction removes features that do not preserve useful
target structure. GenZ therefore produces readable, reusable semantic
hypotheses paired with quantitative evidence, but it neither models the
population $p(x)$ nor renders new artifacts.

fLSA \citep{xu2025flsa}, by contrast, targets $p(x)$ through a
reconstruction-like likelihood. It represents each document segment with one
of a fixed number of natural-language tags and alternates between assigning
segments to tags and rewriting each tag from its assigned segments. After
learning, fLSA fits a bigram model over tags, samples a tag sequence as an
outline, and renders text conditioned on that outline.

\method\ differs first in the statistical object it learns. fLSA models a
sequence of one-of-$K$ segment tags, while GenZ uses multiple binary semantic
features to predict $y$. \method\ instead represents each artifact by a
compositional attribute configuration, fits a joint distribution over
configurations, and renders $x$ from sampled configurations.
More importantly, \method\ learns these semantic coordinates differently.
fLSA iteratively refines segment-level tags through assignment and summarization. In contrast, we use direct comparisons among artifacts
within small cohorts to propose the semantic coordinates in which
the population will be modeled. Our objective is not to recover every semantic mode exhaustively, but to expose
meaningful population variation that direct generation systematically
underrepresents.

The semantic contrasts used here have closer relatives in GenZ and D5. GenZ
shows the foundation model two statistically constructed groups whose split is
directed by $y$. D5 compares two predefined corpora and asks for
natural-language differences between them \citep{zhong2023d5}. In \method, the
cohorts are unpartitioned: the foundation model enumerates all attributes
that recur or vary among their members. Near-duplicate proposals are
consolidated, and every retained attribute is back-applied to the full corpus.
Local comparisons therefore propose the vocabulary, while global labeling
determines its empirical frequencies and dependencies. LLooM, TopicGPT, and
TnT-LLM similarly induce natural-language concepts or taxonomies and assign
them back to a corpus, while natural-language-parameterized statistical models
optimize such predicates for clustering, classification, or time-series
objectives \citep{lam2024lloom,pham2024topicgpt,wan2024tnt,zhong2024explaining};
these methods analyze observed data rather than sampling a joint configuration
distribution for rendering.
\paragraph{Sampling sparse attribute configurations.}
Once artifacts are labeled, sampling becomes a classical discrete density-estimation problem. Resampling observed configurations preserves the empirical joint but cannot form new combinations; independent marginals do so while discarding all dependencies. Chow--Liu trees provide a tractable middle ground by retaining the strongest pairwise mutual-information dependencies \citep{chow1968dependence}, while count backoff in language models retains specific contexts only when they have adequate support \citep{katz1987backoff}. Our sampler is a pragmatic higher-order variant: under a fresh random variable ordering, it ranks previously sampled attributes by mutual information with the next attribute, retains the largest sufficiently supported conditioning set, smooths the resulting conditional, and masks incompatible values. We do not claim ancestral sampling, mutual information, or backoff as new; the contribution is their combination for sparse, corpus-induced semantic configurations.
\paragraph{Structured conditional generation.}
Hierarchical text generation separates plans or outlines from prose \citep{fan2018hierarchical,yao2019plan,xu2025flsa}, while scene-layout models separate explicit object and spatial structure from pixel rendering. LayoutVAE and LayoutDM additionally learn distributions over structured layouts before rendering \citep{jyothi2019layoutvae,inoue2023layoutdm,feng2023layoutgpt}. Work such as Design2Code demonstrates increasingly capable screenshot-conditioned webpage rendering \citep{si2025design2code}, but optimizes the fidelity of one page rather than diversity among pages generated for the same request. These works validate separating structure from realization; \method\ instead induces the semantic control space itself from a corpus and samples from a corpus-estimated distribution over attribute configurations before invoking a replaceable conditional renderer.
\paragraph{Population-level evaluation.}
Recent work finds that individually plausible LLM user simulators remain distributionally different from real users and are often overly cooperative or behaviorally uniform \citep{mehri2026measuring,zhou2026sim2real}. When a target population is observed, divergence and coverage measure whether generated frequencies match it; for open-ended sets without a unique target distribution, the Vendi Score provides a reference-free effective-diversity measure, while quality-filtered semantic diversity excludes varied but invalid outputs \citep{friedman2023vendi,shypula2025effective}. These distinctions motivate evaluating distributional fidelity, set diversity, condition adherence, and per-output acceptability separately.

\section{Implementation Details}

\subsection{Ancestral Sampling}
\label{app:ancestral-sampling}
With conditioning on the user request implicit, we draw a random attribute ordering $\pi$ and sample each attribute $a_{\pi_j}$ sequentially using its conditional probability given preceding assignments, $p(a_{\pi_j} \mid a_{\pi_1}, \ldots, a_{\pi_{j-1}})$. However, as $j$ increases, exact empirical conditionals suffer from data sparsity; strictly enforcing them would restrict the sampler to reproducing only exact combinations present in the source dataset. To enable generalization beyond the training data, we dynamically reduce the conditioning set: we rank the previously sampled attributes by their mutual information with $a_{\pi_j}$ and retain only the highest-ranked subset such that a minimum support threshold of $m$ source rows remains. Letting $S$ be this final supported set of source rows, we estimate the conditional probability using additive smoothing:

$$
\widehat{q}(a_{\pi_j} = v \mid S) = \frac{n_v(S) + \beta}{|S| + \beta K_j},
$$

where $n_v(S)$ counts the rows in $S$ with value $v$, $K_j$ is the total number of possible values, and $\beta$ is the smoothing parameter. Finally, to ensure logical coherence, we enforce a pairwise compatibility mask generated offline by an LLM, which reviews schema descriptions to identify mutually exclusive property pairs. During sampling, any candidate value that conflicts with a previously selected attribute is assigned zero probability, and the remaining valid options are renormalized.

\subsection{Weather Reports}
\label{app:weather-mining}

\subsubsection{Dataset Construction}

This controlled experiment uses synthetic text grounded in real weather observations, rather than independently sampled weather attributes. We obtain daily historical weather from the Open-Meteo archive for Phoenix, Reykjavik, Singapore, London, Seattle, Cairo, Moscow, Mumbai, Sydney, and Denver. The training year is 2016, with 366 days per city and 3,660 city-day records. The held-out year is 2017, with 365 days per city and 3,650 records. We apply the same discretization to both years along four categorical weather axes, each with four possible values, represented as 16 binary attributes with one active value per axis. Ten additional one-hot city indicators give 26 binary attributes in the complete source record. City is a conditioning variable throughout evaluation, not one of the weather axes whose joint distribution is scored.

\paragraph{Temperature.}
Daily mean temperature is binned as cold below $8^\circ\mathrm{C}$, mild from $8^\circ\mathrm{C}$ to below $18^\circ\mathrm{C}$, warm from $18^\circ\mathrm{C}$ to below $27^\circ\mathrm{C}$, and hot at or above $27^\circ\mathrm{C}$.

\paragraph{Sky.}
Weather codes 45 and 48 take precedence and map to foggy. Otherwise, mean cloud cover below 25\% maps to clear, from 25\% to below 60\% to partly cloudy, and at least 60\% to overcast.

\paragraph{Wind.}
Gusty takes precedence when the daily maximum gust is at least 60\,km/h or the maximum 10\,m wind speed is at least 40\,km/h. Otherwise, maximum wind speed at least 25\,km/h maps to windy, at least 12\,km/h to breezy, and below 12\,km/h to calm. Thus, windy and gusty are distinct source categories.

\paragraph{Precipitation.}
Weather codes 95, 96, and 99 map to storm. Otherwise, snowfall greater than 0.1\,cm or a snow code in $\{71,73,75,77,85,86\}$ maps to snow. Among the remaining days, precipitation at least 1\,mm or a rain code in $\{51,53,55,56,57,61,63,65,66,67,80,81,82\}$ maps to rain; all other days map to none. The precedence is storm, then snow, then rain, then none. In preprocessing, days with missing mean temperature are skipped; missing cloud cover defaults to 50\%, and missing wind, gust, precipitation, snowfall, or weather-code values are treated as zero.

\begin{table}[h]
    \centering
    \small
    \setlength{\tabcolsep}{3pt}
    \caption{Data-generating and recovered weather schemas. Names are the exact retained identifiers; the correspondence is a semantic alignment, not an exact string match. Source support counts the intended configurations of the 300 rendered source reports, not labels extracted from their text. A dash means no distinct corresponding attribute was recovered. The ten city indicators are supplied metadata and are not included in these 16 weather rows.}
    \label{tab:weather-schema-recovery}
    \begin{tabular}{@{}lp{0.27\linewidth}p{0.34\linewidth}r@{}}
        \hline
        Axis & Data-generating attribute & Recovered attribute & \shortstack{Source\\support} \\
        \hline
        Temperature & \texttt{temp\_cold} & \texttt{temperature\_cold} & 64 \\
        & \texttt{temp\_mild} & \texttt{temperature\_mild} & 97 \\
        & \texttt{temp\_warm} & \texttt{temperature\_warm} & 99 \\
        & \texttt{temp\_hot} & \texttt{temperature\_hot} & 40 \\
        \hline
        Sky & \texttt{sky\_clear} & \texttt{sky\_clear} & 74 \\
        & \texttt{sky\_partly} & \texttt{sky\_partly\_cloudy} & 84 \\
        & \texttt{sky\_overcast} & \texttt{sky\_overcast} & 142 \\
        & \texttt{sky\_foggy} & --- & 0 \\
        \hline
        Wind & \texttt{wind\_calm} & \texttt{wind\_calm} & 71 \\
        & \texttt{wind\_breezy} & \texttt{wind\_breezy} & 194 \\
        & \texttt{wind\_windy} & --- & 24 \\
        & \texttt{wind\_gusty} & \texttt{wind\_gusty} & 11 \\
        \hline
        Precipitation & \texttt{precip\_none} & \texttt{precip\_none} & 159 \\
        & \texttt{precip\_rain} & \texttt{precip\_rain} & 121 \\
        & \texttt{precip\_snow} & \texttt{precip\_snow} & 20 \\
        & \texttt{precip\_storm} & --- & 0 \\
        \hline
    \end{tabular}
\end{table}

We select 30 distinct 2016 days per city and render their configurations with GPT-5-mini, producing 300 source reports. Each request provides the city and four terse weather values, requires all conditions to be expressed in natural wording, and explicitly forbids replacing an unusual configuration with a more plausible one. The output is a city-naming headline followed by a one- to three-sentence paragraph. A private entropy token and instructions to vary voice, length, vocabulary, rhythm, and punctuation encourage surface variation without changing the supplied conditions.

\subsubsection{Recovered Schema and Annotation}

The retained discovery request compares 299 source reports together in a single call; one report is held out from vocabulary discovery. The prompt asks what is shared and what differs, then requests binary, atomic properties named by a category prefix and realization suffix, with a supporting quotation. It does not receive the numerical discretization rules or the predefined property menu. This produces 13 weather attributes across temperature, sky, wind, and precipitation. City is supplied separately rather than discovered as a weather attribute.

Table~\ref{tab:weather-schema-recovery} shows both successful recovery and its limits. The recovered schema distinguishes all four temperature values and three values each for sky, wind, and precipitation. Foggy and storm have zero support in the source configurations, and neither occurs in the retained full-year tuples for 2016 or 2017. By contrast, windy occurs in 24 source configurations but has no separate recovered value. Its absence cannot be explained by zero source support, although this comparison alone cannot distinguish a source-rendering failure from a mining failure.

\subsubsection{City-Conditional Attribute Sampling}

For \method{}, we fit the ancestral sampler to the 300 completed recovered-schema assignments. For each draw, city is clamped and always conditions the source-row selection; it is never dropped during backoff. The four free weather axes are sampled in a fresh random order. Previously sampled free axes are considered as conditioners in descending pairwise mutual-information order, and a conditioner is retained only if at least $m=10$ matching source rows remain. We use additive smoothing $\beta=1$, adding one to each candidate value count before normalization. Each city therefore starts from its 30 source reports, with more detailed conditioning retained only when the support floor permits it.

A GPT-5-mini compatibility pass assesses cross-axis value pairs. The retained induced-schema mask rejects snow with hot or warm temperatures. Candidate values incompatible with already assigned values receive zero weight before normalization when at least one candidate remains allowed; if all values would be rejected, the implemented sampler leaves the unmasked weights in place rather than producing an empty distribution. These are model-judged compatibility constraints, not guarantees of physical impossibility.

We draw 300 configurations per city in the fixed order Phoenix, Reykjavik, Singapore, London, Seattle, Cairo, Moscow, Mumbai, Sydney, Denver. Draws are independent forward passes.

\subsubsection{Generation Conditions and Valid Outputs}

All four reported conditions use GPT-5-mini for text generation and subsequent attribute annotation. The prompt-only baseline receives the city and asks the generator to choose realistic conditions. The append baseline supplies the complete 300-report source collection, including city labels, and requests a fresh report from the same city-conditional distribution rather than a copied example or a stereotypical day. For \method{}, the generator receives the city and the sampled recovered temperature, sky, wind, and precipitation values, with an explicit instruction to preserve unusual combinations rather than silently correct them.

Each condition generates 300 reports per city, or 3,000 overall. All authoring formats request a headline and a one- to three-sentence body, with a private style seed that must not be printed. 

The annotator receives the generated report and the original 16-attribute schema $\mathcal{A}_{\text{source}}$. It attempts to recover one value for each of the four original axes. Reports without a complete extracted tuple after the permitted attempts are omitted from the distributional score, rather than being assigned their intended configuration. Table~\ref{tab:weather-evaluation-counts} gives the resulting sample sizes. 

\begin{table}[h]
    \centering
    \small
    \setlength{\tabcolsep}{4pt}
    \caption{Weather reports generated and retained for distributional evaluation. All conditions generate 300 reports per city. Prompt-only exclusions are incomplete four-axis annotations after retries, not ungenerated reports; its retained counts range from 244 to 300 per city.}
    \label{tab:weather-evaluation-counts}
    \begin{tabular}{@{}lrrr@{}}
        \hline
        Condition & Generated & Complete labels & Excluded \\
        \hline
        Prompt only & 3,000 & 2,838 & 162 \\
        Append reports & 3,000 & 3,000 & 0 \\
        \method{} recovered schema & 3,000 & 3,000 & 0 \\
        Known-schema control & 3,000 & 3,000 & 0 \\
        \hline
    \end{tabular}
\end{table}

\subsubsection{Held-Out Distributional Evaluation}

For each city, the reference distribution is the empirical joint distribution of discretized temperature, sky, wind, and precipitation in 2017. The candidate distribution is estimated from the original-schema tuples extracted from generated text. Thus, scoring measures what the generated reports express, not merely what the sampler requested. We compute each metric separately by city and take the unweighted arithmetic mean over the ten cities.

For a city, let $p$ and $q$ be normalized reference and generated frequencies over the union of their occupied joint cells, with $K$ cells. The forward-KL implementation adds $1/K$ to each \emph{normalized probability}, then renormalizes:
\begin{equation}
    \widetilde p_i=\frac{p_i+1/K}{2},\qquad
    \widetilde q_i=\frac{q_i+1/K}{2},\qquad
    D_{\mathrm{KL}}=\sum_{i=1}^{K}\widetilde p_i
    \log\frac{\widetilde p_i}{\widetilde q_i}.
    \label{eq:weather-forward-kl}
\end{equation}
This is a half-uniform mixture, not add-one smoothing of raw report counts. Jensen--Shannon divergence uses the \emph{unsmoothed} $p$ and $q$, with midpoint $(p+q)/2$ and zero-probability terms omitted. Both divergences use natural logarithms. Coverage is the fraction of reference-occupied joint cells also occupied by generated outputs, without probability-mass weighting; it is not the fraction of reports labeled successfully.

\subsection{User Simulation}
\label{app:user-simulation}

\subsubsection{Evaluation and Schema-Construction Data}

We use the 5,000-conversation coding split of the user-behavior benchmark by \cite{mehri2026measuring} as the real evaluation population. The released real-matched-goal file pairs every evaluation conversation with a different held-out WildChat conversation selected by the benchmark authors using goal-embedding similarity. To keep schema construction disjoint from the 5,000 reference conversations, we use a held-out set of conversations as the candidate schema-training population. We remove candidates whose normalized content duplicates an evaluation conversation and conversations with at least 30,000 characters.

GPT-5-mini with low reasoning effort classifies the language of the user's natural-language instructions while ignoring source code, logs, and quoted artifacts. Of 3,585 English-eligible conversations, we sample 600 without replacement using seed 42. Both the compact and fine-grained schemas are learned from this same frozen sample.

\paragraph{User goal reconstruction.}
The public artifact consumed by our pipeline contains only \texttt{id}, \texttt{real\_messages}, and \texttt{simulated\_messages}; it does not contain the private per-conversation goal field used by the original simulator. We therefore reconstruct one \emph{goal} per evaluation conversation with GPT-5.4 at low reasoning effort. The reconstruction prompt requests a one- or two-sentence, approximately 40-word statement of the high-level objective. It explicitly forbids copied code, literal text or data, variable names, exact formatting requirements, fine-grained constraints, and descriptions of user behavior. This prevents the simulator from receiving the real user's artifact or interaction strategy before the conversation begins. The resulting 5,000 goals are frozen, and the same goal for an identifier is reused for every model and method.

\subsubsection{Schema Mining and Consolidation}

The miner compares cohorts of real conversations and proposes topic-agnostic, binary behavioral attributes grounded in exact user quotations. Attributes describe how users make requests, supply context, react to responses, specify desired outputs, and manage the interaction rather than what their coding task is about. For each recipe, the initial 60 conversations are partitioned into five cohorts of 12, verified from the saved mining manifests. Cohort packing permits at most 12 conversations and 28,000 tokens per request; assistant turns are truncated to 400 characters while user text is retained. GPT-5-mini uses low reasoning effort and a 16,384-token completion budget for seed mining.

Seed consolidation first canonicalizes the emergent axis names, then merges same-meaning realizations within each canonical axis using their supporting user quotations. In the compact recipe, axes jointly assigned to a conversation are explicitly constrained to remain separate during axis canonicalization. We then incorporate the remaining 540 conversations in nine update rounds of 60. Each conversation is processed in its own request containing the current schema: GPT-5-mini labels supported existing attributes, proposes genuinely novel behavior, and provides a user-turn index and quotation for each label. At the end of each round, a separate consolidation pass folds the collected novel proposals into existing or new axes. Thus, a 60-conversation update round is not one 60-conversation mining request. The final vocabulary is back-applied to all 600 conversations. A separate GPT-5-mini pass produces a compatibility mask for mutually incompatible attribute values.

We retain two mining recipes. The \emph{compact} recipe uses categorical axis mining, requires a novel realization to have support from at least two conversations, and uses strict multi-value resolution. It produces 40 axes and 84 attributes. The \emph{fine-grained} recipe uses permissive mining, retains novel realizations with support of one, and resolves multi-value axis assignments by their most frequent realization. It produces 292 axes and 465 attributes.

\subsubsection{Attribute Frequencies and Co-Occurrence}

Tables~\ref{tab:user-attributes} and~\ref{tab:user-cooccurrences} characterize the empirical attribute matrices used for sampling. We count each non-absent attribute once per source conversation, with a denominator of 600. We count unordered co-occurring pairs only across different axes. Attributes and pairs are ranked separately within each schema by decreasing support, with ties broken alphabetically by the exact frozen identifiers.

\begin{table}[h]
    \centering
    \footnotesize
    \setlength{\tabcolsep}{3pt}
    \caption{The five most frequent non-absent attributes per user schema after back-applying the frozen vocabulary to all 600 training conversations. Identifiers are retained verbatim.}
    \label{tab:user-attributes}
    \begin{tabular}{@{}p{0.16\linewidth}p{0.52\linewidth}rr@{}}
        \hline
        Schema & Attribute & Count & Frequency \\
        \hline
        Compact & \texttt{tone\_\allowbreak no\_\allowbreak greeting\_\allowbreak direct} & 539 & 89.83\% \\
        Compact & \texttt{expectation\_\allowbreak of\_\allowbreak code\_\allowbreak full\_\allowbreak snippet} & 415 & 69.17\% \\
        Compact & \texttt{request\_\allowbreak imperative\_\allowbreak command} & 366 & 61.00\% \\
        Compact & \texttt{no\_\allowbreak or\_\allowbreak small\_\allowbreak code\_\allowbreak sample} & 283 & 47.17\% \\
        Compact & \texttt{context\_\allowbreak provides\_\allowbreak short\_\allowbreak example\_\allowbreak or\_\allowbreak none} & 274 & 45.67\% \\
        \hline
        Fine-grained & \texttt{request\_\allowbreak imperative\_\allowbreak command} & 409 & 68.17\% \\
        Fine-grained & \texttt{wants\_\allowbreak full\_\allowbreak runnable\_\allowbreak code\_\allowbreak or\_\allowbreak complete\_\allowbreak artifact} & 316 & 52.67\% \\
        Fine-grained & \texttt{provides\_\allowbreak code\_\allowbreak or\_\allowbreak artifact\_\allowbreak example} & 288 & 48.00\% \\
        Fine-grained & \texttt{single\_\allowbreak shot\_\allowbreak request} & 225 & 37.50\% \\
        Fine-grained & \texttt{no\_\allowbreak greeting\_\allowbreak direct} & 208 & 34.67\% \\
        \hline
    \end{tabular}
\end{table}

\begin{table}[h]
    \centering
    \scriptsize
    \setlength{\tabcolsep}{2pt}
    \caption{The three most frequent unordered cross-axis attribute pairs per user schema. Joint frequency is the fraction of the 600 source conversations containing both attributes. Lift divides joint frequency by the product of the two marginal frequencies.}
    \label{tab:user-cooccurrences}
    \begin{tabular}{@{}p{0.12\linewidth}p{0.29\linewidth}p{0.29\linewidth}rr@{}}
        \hline
        Schema & Attribute A & Attribute B & \shortstack{Joint\\count} & Lift \\
        \hline
        Compact & \texttt{expectation\_\allowbreak of\_\allowbreak code\_\allowbreak full\_\allowbreak snippet} & \texttt{tone\_\allowbreak no\_\allowbreak greeting\_\allowbreak direct} & 376 (62.67\%) & 1.009 \\
        Compact & \texttt{request\_\allowbreak imperative\_\allowbreak command} & \texttt{tone\_\allowbreak no\_\allowbreak greeting\_\allowbreak direct} & 343 (57.17\%) & 1.043 \\
        Compact & \texttt{expectation\_\allowbreak of\_\allowbreak code\_\allowbreak full\_\allowbreak snippet} & \texttt{request\_\allowbreak imperative\_\allowbreak command} & 297 (49.50\%) & 1.173 \\
        \hline
        Fine-grained & \texttt{request\_\allowbreak imperative\_\allowbreak command} & \texttt{wants\_\allowbreak full\_\allowbreak runnable\_\allowbreak code\_\allowbreak or\_\allowbreak complete\_\allowbreak artifact} & 258 (43.00\%) & 1.198 \\
        Fine-grained & \texttt{evidence\_\allowbreak pastes\_\allowbreak existing\_\allowbreak code\_\allowbreak snippet} & \texttt{provides\_\allowbreak code\_\allowbreak or\_\allowbreak artifact\_\allowbreak example} & 182 (30.33\%) & 1.841 \\
        Fine-grained & \texttt{provides\_\allowbreak code\_\allowbreak or\_\allowbreak artifact\_\allowbreak example} & \texttt{request\_\allowbreak imperative\_\allowbreak command} & 182 (30.33\%) & 0.927 \\
        \hline
    \end{tabular}
\end{table}

Common pairs need not be strongly associated: several high-support pairs have lift close to one because both constituent attributes are individually common. The tables describe the source behavior distribution, not whether generated conversations realize every requested attribute.

\subsubsection{Ancestral Profile Sampling}

For sampling, each axis is represented by its retained realizations plus an explicit absent value. Realizations with empirical support below eight are removed from the sampling manifold. For each synthetic profile, we draw a fresh random ordering of the axes. When sampling axis $a_c$, previously drawn non-absent axes are ranked by pairwise mutual information with $a_c$. Starting from all 600 rows, we add a previous assignment to the conditioning set only when the resulting intersection retains at least $m=10$ source conversations. If $S$ is the retained row set and axis $a_c$ has $K_c$ possible values, its pre-mask probability is
\begin{equation}
    P(a_c=v\mid S)=\frac{N_S(a_c=v)+\beta}{|S|+\beta K_c},
    \qquad \beta=0.2.
    \label{eq:user-ancestral}
\end{equation}
Values forbidden by the compatibility mask receive zero probability and the remaining probabilities are renormalized. We sample 5,000 profiles from each schema with seed 42. The compact sampler uses strict multi-axis resolution and the fine-grained sampler uses most-frequent resolution. These same frozen profile files are reused for every user model so that model comparisons do not resample the conditioning distribution.

\subsubsection{Simulation Arms and Conversation Harness}

Every user model is evaluated under four matched simulation conditions. \emph{Prompt only} supplies the frozen thin goal and generic instructions to behave realistically, split information across turns, and terminate when the goal is satisfied. \emph{PersonaHub} adds one persona sampled uniformly with replacement from a frozen pool of 5,855 coding-relevant personas, identified by filtering the collection of personas with GPT-5.6 Sol. The 5,000 persona assignments are generated with seed 42 and reused across user models. The prompt makes the goal authoritative: the persona can affect knowledge, vocabulary, assumptions, and interaction style but cannot replace the task. \method{} \emph{compact} and \method{} \emph{fine-grained} add the corresponding sampled behavioral profile. All non-absent sampled attributes are included. When available, each attribute is grounded with one real-user quotation truncated to 140 characters; the prompt instructs the simulator to imitate the behavior rather than copy the quotation.

The assistant is fixed to GPT-5.4 with low reasoning effort in every condition. It begins with ``How can I help you?''; this seed turn is removed from the saved transcript. To generate a user turn, the simulator receives its system prompt and a role-reversed history, so prior assistant messages appear as user messages in the simulator's local context and vice versa. The generated response becomes the next public user message. The assistant receives the ordinary, non-reversed public history. Generation stops when the simulator emits \texttt{<|TERMINATE\_CONVERSATION|>}, after ten simulated user turns, or before the transcript would exceed 30,000 characters. User and assistant calls allow at most 4,096 completion tokens.

GPT-5-mini and GPT-5.5 use low reasoning effort. GPT-4o and Kimi K2.6 are run without an explicit reasoning-effort control. Qwen3.8-27B runs on two NVIDIA A100 80GB GPUs with tensor parallelism of two and a 262,144-token context window. Qwen thinking is enabled, but private reasoning state is retained only in memory between simulator turns and removed before transcripts are saved, sent to the GPT-5.4 annotator, or published.

\subsubsection{Matched Evaluation Sets}

For some user goals, different model endpoints will block the generation due to content policy or return empty messages. For each user model, we independently form the common identifier intersection of the real, prompt-only, PersonaHub, compact, and fine-grained arms. An identifier missing from any arm, containing no non-empty user message, blocked by provider policy, or failing behavioral annotation after retries is removed from all five arms for that model. This preserves paired comparisons while allowing different providers to have different policy exclusions. Table~\ref{tab:user-model-sizes} gives the final effective sample sizes.

\begin{table}[h]
    \centering
    \small
    \setlength{\tabcolsep}{4pt}
    \caption{User simulators and final matched evaluation sizes.}
    \label{tab:user-model-sizes}
    \begin{tabular}{@{}p{0.25\linewidth}p{0.47\linewidth}r@{}}
        \hline
        User simulator & Inference control & Effective $N$ \\
        \hline
        GPT-5-mini & Low reasoning & 4,960 \\
        GPT-5.5 & Low reasoning & 4,903 \\
        GPT-4o & No explicit reasoning control & 4,946 \\
        Kimi K2.6 & No explicit reasoning control & 4,935 \\
        Qwen3.8-27B & Thinking enabled; private state removed & 4,956 \\
        \hline
    \end{tabular}
\end{table}

\subsubsection{Evaluation}
We follow the exact evaluation protocol and prompts of \cite{mehri2026measuring}. We use GPT-5.4 with low reasoning effort as an annotator for the six facets, and text-embedding-3-large as the embedding model, keeping the first 1024 embedding dimensions.

\subsection{Webpage Generation}
\label{app:webpage-generation}

\subsubsection{Tasks and Experimental Conditions}

We evaluate webpage generation on 100 synthetic authoring tasks paired with grounding documents containing the desired content. The source documents comprise 79 DOCX files and 21 PDFs; 79 tasks contain a total of 166 extracted images. For each task, we generate ten pages under each of three conditions: prompt-only generation, \method{} using the WebSight schema, and \method{} using the Design2Code schema. We repeat this experiment with GPT-5.5 and Kimi K2.6, yielding 3,000 accepted pages per generator and 6,000 pages overall. All three conditions receive the same authoring request, extracted grounding text, and available source images. Images are converted to PNG, supplied as visual model inputs with their local asset paths and descriptions, and made available to the generated page. Previously generated webpages and their screenshots are not supplied as inputs. The prompt-only condition uses the same generation template but omits the sampled design-system block.

\subsubsection{Schema Mining and Consolidation}

\paragraph{WebSight.}
We discover the vocabulary from 1,000 webpages partitioned into 100 disjoint cohorts of ten. Each mining request supplies the cohort's HTML and screenshots to GPT-5-mini and asks for important, transferable binary design properties, including both shared traits and differences. The output records a property name, aspect, description, and the pages exhibiting it. The clean consolidation starts from 899 raw property names and their observed descriptions, not from an earlier consolidated vocabulary.

For consolidation, we embed each property with its descriptions using \texttt{text-embedding-3-large} with 1,024 retained dimensions and retrieve up to 60 neighbors at cosine similarity at least 0.80. The resulting 9,889 candidate pairs are judged with a strict bidirectional-entailment prompt: equivalence requires that every page satisfying either property also satisfies the other. Tentative matches receive an adversarial verification pass. Complete-link grouping requires every pair of members to be accepted as equivalent; the canonical label is an existing high-frequency member rather than a newly generalized label. This yields 485 attributes. We back-apply the complete canonical vocabulary to each source page using its HTML, screenshot, and existing mining labels, preserving those labels and adding supported missing attributes. We subsequently annotate 9,000 additional pages with the same frozen vocabulary, yielding the 10,000-row sampling matrix.

\paragraph{Design2Code.}
The final 5,484-attribute vocabulary is mined with GPT-5.6-sol at high reasoning effort from 50 five-page cohorts, comprising 250 pages. Each request supplies compact HTML and screenshots. The prompt first inventories consequential decisions on each page, then harmonizes equivalent names across the cohort while preserving component identity, atomicity, and transferable meaning. A property may be retained even if it occurs on only one page in its cohort. This differs from simply requesting the most salient shared traits.

The retained schema records 7,161 mined property records and 5,906 unique names. Name-based filtering and hierarchical routing retain 5,675 names and assign component owners and design-axis families. The hierarchy supports retrieval and annotation; the sampling variables remain flat binary leaf properties. Candidate synonyms are retrieved locally within component-family/design-axis groups, with additional family-wide and global candidates. Name embeddings use 1,024 dimensions: local retrieval uses up to 30 neighbors at cosine similarity at least 0.78, family retrieval ten at 0.84, and global retrieval six at 0.88; exact token-signature and hierarchy-coordinate matches provide additional candidates. GPT-5-mini judges name-only equivalence, GPT-5.6-sol adversarially verifies proposed matches, and complete-link grouping produces the 5,484 canonical attributes.

The fixed vocabulary is then annotated across 480 pages selected from the 484-page benchmark, rather than only the 250 mining pages. Annotation uses two GPT-5.6-sol calls per page: a high-recall selector identifies relevant component-family/design-axis routes, including uncertain routes, and a second call assigns supported canonical names from those routes using HTML and screenshot evidence. This assignment starts from an empty vector rather than trusting the original mining labels. The retained hierarchy contains 42 component families and 669 family/axis routes. Accordingly, neither 484 nor 480 is the vocabulary-discovery sample size for this schema.

\subsubsection{Schema Sampling and Request-Specified Attributes}

The WebSight condition uses a frozen schema of 485 binary design attributes with an empirical attribute matrix covering 10,000 webpages. This matrix extends an existing 1,000-page matrix by annotating an additional 9,000 pages with the same schema; the extension does not introduce new attributes. The Design2Code condition uses the 5,484-attribute schema and its empirical matrix over 480 webpages selected from the 484-page benchmark. Both conditions use their saved pairwise incompatibility masks.

For each task and schema, GPT-5.5 identifies attributes explicitly specified by the authoring request, using high reasoning effort and a 16,384-token completion budget. The derivation input contains the authoring request and schema menu, rather than the grounding document. Invalid structured outputs receive at most two additional attempts with a JSON-format correction. Unknown property names are rejected; conflicting specified properties are resolved in their returned order, retaining the first compatible properties and recording those discarded.

We use the ancestral sampler described in Appendix~\ref{app:ancestral-sampling}, with additive smoothing $\beta=1.0$ for WebSight and $\beta=0.01$ for Design2Code, and a minimum conditioning support of $m=10$ source pages for both. Previous attribute assignments are considered in descending mutual-information order, retaining a conditioner only when at least ten compatible source rows remain. Importantly, request-specified attributes are enforced \emph{after} a draw from the full empirical matrix: sampled attributes incompatible with a specified attribute are removed, and the specified attributes are then added. We do not first restrict the empirical matrix to rows satisfying the request, since rare specified attributes could collapse the available support.

We save ten design draws per task and schema. Each design is provided to the generator as a JSON list of property names and descriptions, with an instruction to realize all listed properties. The Kimi experiment reuses exactly the same 200 task-schema caches, including all ten designs and the GPT-5.5-derived specified attributes. Thus, the generators receive matched sampled designs; attribute derivation is not independently repeated with Kimi.

\subsubsection{Generation, Validity Checks, and Recovery}

Generation uses GPT-5.5 with low reasoning effort and a 16,384-token completion budget, or Kimi K2.6 with a 32,768-token output budget and default reasoning.

The generation template requests a complete, self-contained HTML/CSS page. Pages may reference supplied local images or validated self-contained SVG data images, but may not load external or missing resources. Static validation checks HTML completeness, rejects scripts and inline event handlers, and validates resource references and embedded SVG content. Kimi receives an additional explicit clarification prohibiting scripts, JavaScript URLs, and event handlers, including nominally harmless form handlers; static HTML/CSS equivalents take precedence over requested interactivity.

\subsubsection{Browser Rendering}

We render accepted pages with Playwright Chromium at a $1280\times768$ viewport and device scale factor one. Page JavaScript and service workers are disabled, external HTTP(S) requests are blocked, and animations and transitions are suppressed. We wait for fonts and image decoding, force image loading to be eager, and reject pages with fewer than 100 characters of rendered body text or any broken images. These checks establish basic renderability, not content fidelity or usability.

For a page of height $H$, we capture viewport screenshots at vertical positions zero, the rounded midpoint of the scroll range, and the bottom of the scroll range, where the maximum scroll is $\max(0,H-768)$. Each capture follows a 100-ms settling interval. Short pages can therefore have identical views. We retain all three screenshots, giving 9,000 screenshots per generator. Odd-one-out evaluation uses only the top view; visual diversity uses all three.

\subsubsection{Odd-One-Out Evaluation}

For each task and generator, we separately compare each schema-conditioned set against the prompt-only set. Each comparison begins with a pool of ten baseline and ten schema-conditioned top-view screenshots. Screenshots receive neutral numeric labels, and the judge is asked to identify the single page whose visual design is most unlike the others, considering layout, colour, typography, spacing, and components. The judge receives screenshots, not HTML, schema identifiers, or condition labels.

We use GPT-5-mini with low reasoning effort and a 2,048-token completion budget. Each response must contain a valid JSON object identifying an in-pool label and a short reason; malformed responses receive up to two additional parsing attempts rather than a substituted selection. After each decision, we remove the selected page and reshuffle and relabel the remaining pool. We stop after ten removals and repeat the complete cascade four times. Shuffle seeds are derived deterministically from the experiment seed, task, schema condition, repetition, and round. This yields 200 task-schema comparisons and 8,000 removal decisions per generator.

We report the fraction of removals attributable to the schema-conditioned set, with 4,000 removals as the denominator for each schema-generator pair. We also report mean removal rank: the ten removed pages receive ranks one through ten in removal order, while \emph{every surviving page receives rank 20}, not an inferred rank between 11 and 20. Ranks are averaged across repetitions and pages within a condition, then equally across tasks. The baseline rank is comparison-specific because it depends on the competing schema-conditioned pool. Larger removal shares and smaller ranks indicate greater stylistic distinctiveness in these mixed pools; neither is a quality-preference or task-success measure.

\subsubsection{Vendi Score evaluation}

\paragraph{Structural representation.}
For structural diversity, we extract a copy-stripped representation of the rendered DOM and computed CSS. Visible elements are described in DOM order using semantic tags or roles, nesting depth, layout and positioning, discretized geometry, typography, foreground and background colour categories, borders, corner radii, shadows, child counts, and text-length bins. Flex, grid, media, and form elements receive corresponding structural descriptors. The document summary includes height class, horizontal overflow, component counts, and a semantic-tag sequence. Literal copy, CSS class names, asset URLs, and image alt text are omitted; text amount is retained only through coarse bins. If necessary, element descriptions are subsampled evenly across DOM order to fit a 7,800-token limit measured with \texttt{cl100k\_base}. A \texttt{text-embedding-3-large} model embeds each resulting document into 1,024 dimensions.

\paragraph{Visual representation.}
For visual diversity, we use DINOv2-G. Each screenshot is converted to RGB and resized bicubically from $1280\times768$ to $560\times336$ without cropping. Each view vector is L2-normalized; the top, middle, and bottom vectors are concatenated in that order and normalized again, producing a 4,608-dimensional page representation.

\paragraph{Vendi Score Computation}

We L2-normalize page embeddings and compute Vendi scores separately for each task and condition using its ten pages using the RBF kernel. For unit-normalized embeddings $x_i$ and $x_j$, the kernel is
\begin{equation}
    K_{ij}=\exp\left(-\frac{\left\|x_i-x_j\right\|_2^2}{2\sigma^2}\right).
    \label{eq:webpage-rbf}
\end{equation}

\subsubsection{Attribute Frequencies and Co-Occurrence}

Tables~\ref{tab:webpage-attributes} and~\ref{tab:webpage-cooccurrences} characterize the empirical attribute matrices used for sampling. We count binary presence once per source webpage, with denominators of 10,000 for WebSight and 480 for Design2Code. Attributes and unordered pairs are ranked separately within each schema by decreasing support, breaking ties alphabetically by the exact schema identifiers. These are source-page annotations, not measurements of whether a generated page realizes its requested attributes.

\begin{table}[h]
    \centering
    \small
    \setlength{\tabcolsep}{4pt}
    \caption{The five most frequent attributes per schema in the empirical source-page matrices. Identifiers are retained verbatim. Counts use 10,000 WebSight pages and 480 Design2Code pages.}
    \label{tab:webpage-attributes}
    \begin{tabular}{@{}p{0.17\linewidth}p{0.48\linewidth}rr@{}}
        \hline
        Schema & Attribute & Count & Frequency \\
        \hline
        WebSight & \texttt{type\_\allowbreak grotesque\_\allowbreak sans} & 8,134 & 81.34\% \\
        WebSight & \texttt{color\_\allowbreak light\_\allowbreak neutral\_\allowbreak background} & 6,653 & 66.53\% \\
        WebSight & \texttt{header\_\allowbreak present} & 6,192 & 61.92\% \\
        WebSight & \texttt{content\_\allowbreak left\_\allowbreak aligned} & 5,811 & 58.11\% \\
        WebSight & \texttt{top\_\allowbreak h1\_\allowbreak heading\_\allowbreak present} & 4,314 & 43.14\% \\
        \hline
        Design2Code & \texttt{site\_\allowbreak header\_\allowbreak present} & 418 & 87.08\% \\
        Design2Code & \texttt{site\_\allowbreak footer\_\allowbreak present} & 401 & 83.54\% \\
        Design2Code & \texttt{body\_\allowbreak typography\_\allowbreak sans\_\allowbreak serif} & 374 & 77.92\% \\
        Design2Code & \texttt{primary\_\allowbreak navigation\_\allowbreak present} & 346 & 72.08\% \\
        Design2Code & \texttt{page\_\allowbreak layout\_\allowbreak responsive} & 342 & 71.25\% \\
        \hline
    \end{tabular}
\end{table}

\begin{table}[h]
    \centering
    \footnotesize
    \setlength{\tabcolsep}{3pt}
    \caption{The three most frequent unordered co-occurring attribute pairs per schema. Joint frequency is the fraction of source pages containing both attributes. Lift divides joint frequency by the product of the two marginal frequencies; values near one indicate little departure from marginal independence.}
    \label{tab:webpage-cooccurrences}
    \begin{tabular}{@{}p{0.13\linewidth}p{0.25\linewidth}p{0.25\linewidth}rrr@{}}
        \hline
        Schema & Attribute A & Attribute B & \shortstack{Joint\\count} & \shortstack{Joint\\frequency} & Lift \\
        \hline
        WebSight & \texttt{color\_\allowbreak light\_\allowbreak neutral\_\allowbreak background} & \texttt{type\_\allowbreak grotesque\_\allowbreak sans} & 5,462 & 54.62\% & 1.009 \\
        WebSight & \texttt{header\_\allowbreak present} & \texttt{type\_\allowbreak grotesque\_\allowbreak sans} & 5,148 & 51.48\% & 1.022 \\
        WebSight & \texttt{content\_\allowbreak left\_\allowbreak aligned} & \texttt{type\_\allowbreak grotesque\_\allowbreak sans} & 4,821 & 48.21\% & 1.020 \\
        \hline
        Design2Code & \texttt{site\_\allowbreak footer\_\allowbreak present} & \texttt{site\_\allowbreak header\_\allowbreak present} & 372 & 77.50\% & 1.065 \\
        Design2Code & \texttt{primary\_\allowbreak navigation\_\allowbreak present} & \texttt{site\_\allowbreak header\_\allowbreak present} & 337 & 70.21\% & 1.118 \\
        Design2Code & \texttt{body\_\allowbreak typography\_\allowbreak sans\_\allowbreak serif} & \texttt{site\_\allowbreak header\_\allowbreak present} & 325 & 67.71\% & 0.998 \\
        \hline
    \end{tabular}
\end{table}

Common pairs need not be strongly associated: several of the highest-support pairs have lift close to one because both constituent attributes are individually common. These tables therefore describe prevalence rather than evidence of strong dependence. Sampled-design frequencies are a separate quantity, affected by smoothing, incompatibility constraints, and request-specified attributes. For example, \texttt{type\_\allowbreak grotesque\_\allowbreak sans} appears in 719 of the 1,000 saved WebSight designs (71.90\%), while \texttt{body\_\allowbreak typography\_\allowbreak sans\_\allowbreak serif} appears in 809 of the 1,000 Design2Code designs (80.90\%). These design draws are shared by the two generators and are not counted twice.

\paragraph{Highest-lift pairs.}
To complement the prevalence-based examples, we rank distinct unordered attribute pairs  by their empirical lift relative to the product of their marginal probabilities. For an attribute set $S$ of size two, we define
\begin{equation}
    L(S)=\frac{\widehat{P}\bigl(\bigcap_{a\in S}\{a=1\}\bigr)}
                  {\prod_{a\in S}\widehat{P}(a=1)}.
    \label{eq:webpage-itemset-lift}
\end{equation}
Probabilities are estimated from the same source-page matrices as the preceding tables, without smoothing. We require \emph{joint} support of at least 1\% of the source corpus: at least 100 WebSight pages or five Design2Code pages. We rank all qualifying combinations separately by schema and set size, using exact rational lift and alphabetically ordered attribute identifiers to resolve ties. Table~\ref{tab:webpage-high-lift-pairs} shows the first three combinations in each ranking.

\begin{table}[h]
    \centering
    \footnotesize
    \setlength{\tabcolsep}{3pt}
    \caption{Highest-lift attribute pairs with joint support of at least 1\%. The three rows per schema are ranked by lift, not frequency. Several Design2Code pairs tie at lift 96; alphabetical tie-breaking determines the displayed subset.}
    \label{tab:webpage-high-lift-pairs}
    \begin{tabular}{@{}p{0.13\linewidth}p{0.52\linewidth}rrr@{}}
        \hline
        Schema & Attribute pair & \shortstack{Joint\\count} & \shortstack{Joint\\frequency} & Lift \\
        \hline
        WebSight &
        \texttt{buttons\_\allowbreak full\_\allowbreak width},\newline
        \texttt{stacked\_\allowbreak form\_\allowbreak inputs} &
        104 & 1.04\% & 19.10 \\
        WebSight &
        \texttt{header\_\allowbreak fixed},\newline
        \texttt{nav\_\allowbreak fixed\_\allowbreak top} &
        103 & 1.03\% & 11.63 \\
        WebSight &
        \texttt{layout\_\allowbreak sidebar\_\allowbreak offset},\newline
        \texttt{nav\_\allowbreak side\_\allowbreak vertical} &
        161 & 1.61\% & 11.41 \\
        \hline
        Design2Code &
        \texttt{article\_\allowbreak caption\_\allowbreak centered},\newline
        \texttt{article\_\allowbreak image\_\allowbreak captions\_\allowbreak centered} &
        5 & 1.04\% & 96.00 \\
        Design2Code &
        \texttt{body\_\allowbreak copy\_\allowbreak monospace},\newline
        \texttt{body\_\allowbreak typography\_\allowbreak monospace} &
        5 & 1.04\% & 96.00 \\
        Design2Code &
        \texttt{coming\_\allowbreak soon\_\allowbreak content\_\allowbreak upper\_\allowbreak middle},\newline
        \texttt{launch\_\allowbreak status\_\allowbreak uppercase} &
        5 & 1.04\% & 96.00 \\
        \hline
    \end{tabular}
\end{table}

\subsection{Text-to-Image}
\label{app:t2i}

\subsubsection{Mining}
We use GPT-5.6 Sol to mine attributes from 7,879 COCO images. Each cohort contains 5 images from a single category. We then consolidate the attributes with GPT-5.6 Sol to a single schema containing 28,875 attributes. Unknown values are excluded from the relevant probability estimates rather than treated as absences. The topic fit uses fixed feature budgets of 350 mined attributes and 450 lexical features; these are not the total numbers of eligible features and do not restrict the schema available to ancestral sampling.

Eligible mined attributes have at least 20 observed rows, at least 20 positive labels, and 2--95\% prevalence among observed rows. We rank them by $\sqrt{n_+}\log(N/n_+)[1+0.05\min(c,20)]$, where $N=7{,}879$, $n_+$ is positive support, and $c$ is the number of categories associated with the consolidated attribute. We reserve 80 slots for the highest-ranked eligible predicates containing both X and Y, then fill the remaining 270 slots from the ranked, unselected attributes.

Lexical features are extracted from attribute names, aliases, and descriptions using TF--IDF selection of up to four unigrams and five phrases per attribute. For topic fitting, a lexical feature must occur in at least 40 rows and at most 50\% of rows, and must appear in at least two attribute documents unless it is category-derived. Eligible category-derived features are included first; remaining slots are filled by ranking candidates with $\sqrt{n}\log(N/n)[1+0.15(\ell-1)]$, where $n$ is row support and $\ell$ is the number of words in the feature. We fit 20 topics to the resulting $800\times800$ positive-NPMI association matrix. Additional query features are projected into the fixed topic space by nonnegative fitting of their associations with the retained features; projections with positive-association correlation below 0.30 are excluded.

The empirical component constructs a request-specific pool of at most 250 attributes from the full schema, using lexical evidence to reweight observed occurrence probabilities. It uses a support floor of $m=20$ and Beta smoothing with $\beta=0.2$. At each of 20 selection steps, the sampler chooses the empirical or topic component with equal probability and draws an unselected, compatible attribute. Attributes selected by either component condition subsequent empirical draws. Thus the image implementation selects a fixed-size set of positive controls rather than a complete binary assignment to the schema.

\subsubsection{Topic-Ancestral sampling}

Let:

\begin{itemize}
    \item $q$ be the input prompt;
    \item $S$ be the properties already selected;
    \item $q_A(f\mid S,q)$ be the empirical MI-backoff next-property distribution;
    \item $q_T(f\mid q)$ be the prompt-conditioned topic next-property distribution;
    \item $\lambda$ be the topic mixing weight.
\end{itemize}

At each of 20 control-selection steps, the sampler uses:

\[q_\lambda(f\mid S,q) = (1-\lambda)q_A(f\mid S,q) + \lambda q_T(f\mid q).\]

Operationally, each control slot chooses the empirical component with probability $1-\lambda$ or the topic component with probability $\lambda$. The chosen component then samples one compatible property without replacement. Properties selected by either component constrain subsequent empirical draws. Hard compatibility masks apply across the complete configuration.

\paragraph{Empirical ancestral component}

The empirical component first constructs a prompt-specific pool of at most 250 properties. For each property:

\begin{enumerate}
    \item estimate its global positive probability on known cells;
    \item estimate smoothed query-atom likelihood ratios from positive and negative source rows;
    \item weight each atom contribution by $\text{support}/(\text{support}+m)$;
    \item clip one atom's log-odds contribution to $[-2,2]$ and the summed shift to $[-6,6]$;
    \item retain properties with nontrivial prompt evidence.
\end{enumerate}

The experiment uses support floor $m=20$ and Beta smoothing $\beta=0.2$.

During fixed-count sampling, each remaining empirical candidate is conditioned on already selected controls. Previous controls are ranked by binary mutual information with the target. A conditioner is retained only if intersecting its positive rows leaves at least $m$ target-observed rows. The resulting Beta-smoothed probability is combined with the prompt-atom log-odds shift and normalized across feasible remaining properties.

This is a positive-only fixed-count version of the paper's earlier binary MI-backoff ancestral sampler. The separately rendered $\lambda=0$ endpoint uses this same common implementation, so the lambda ablation changes only the mixing weight.

\paragraph{Topic component}

Property loadings are normalized within every topic. Given inferred prompt mixture $\theta(q)$, the topic component forms the marginal next-property distribution:

\[q_T(f\mid q) = \sum_k \theta_k(q) \frac{H_{fk}}{\sum_{f'\in\mathcal F}H_{f'k}},\]

renormalized over unselected, compatible properties. Sampling a property from this marginal is equivalent to first sampling a topic and then sampling a property within that topic.

This is an interpolation between \textbf{next-property distributions}. It is not literal elementwise interpolation of the original TxF matrix with a dense reconstructed matrix.

\subsubsection{Rendering}

The rendering procedure $\mathcal Q$ receives the complete request \(u\) together with the sampled attributes. The request takes priority, and the attributes provide positive, non-exhaustive guidance: unsampled properties remain permissible. Role placeholders are bound independently for each attribute and may refer to entities named in the request or to plausible additional entities. In our image experiments, we compare direct rendering by GPT Image 2, which reconciles the sampled attributes with the prompt during generation, with a two-stage procedure. In the latter, GPT-5.6 Sol is first instructed to preserve the request and the meaning of sampled attributes, seek jointly compatible realizations, and introduce plausible supporting entities when needed. It produces a coherent scene plan and concrete instructions for GPT Image 2, replacing role placeholders with their assigned entities. Both variants permit attributes to be omitted when no physically plausible realization is compatible with the request and the other retained attributes.

Sampled attributes provide a sparse, corpus-derived guide to choices left open by the request. In one recorded example, a request for a close-up of chocolate cake with a fork beside it favored dining-related attributes through corpus-estimated associations. The sampled relation ``X on Y'' led Sol to bind X to the cake and Y to a white plate; GPT Image 2 rendered this support relation, which the request had not specified. Other sampled attributes guided the lighting, warm palette, and dense composition, as well as the instantiation of a nearby mug and a hand holding the plate. Therefore the attributes need not specify a complete scene: they direct the renderer toward particular relevant aspects of reference-corpus variation, leaving it to supply coherent concrete details. Across generations, sampling these controls according to their estimated frequencies and dependencies steers the model's completion of underspecified requests toward the reference corpus at test time.

Both direct and Sol-assisted rendering produce substantially more diverse images than prompt-only and prompt-rewriting baselines, while maintaining high visual plausibility (Section~\ref{t2i}). In our GPT Image 2 comparison ($\lambda=0.5$), direct rendering realizes 61.6\% of sampled attributes. When GPT-5.6 Sol first assigns entities to role placeholders and produces coherent instructions, it manages to reconcile 99.2\% of the sampled attributes, and realization increases to 80.2\% of the originally sampled attributes, according to a post-hoc model-based audit.

\paragraph{Direct variant}

The Direct variant sends the original prompt plus the 20 sampled controls to the image generators. The wrapper states that:

\begin{itemize}
    \item the original prompt has absolute priority;
    \item controls are positive but nonexhaustive;
    \item unspecified properties remain allowed;
    \item the conventional meaning of compound nouns must be preserved;
    \item $X$ denotes the primary requested entity;
    \item $Y$ may be explicit, absent, or plausibly introduced;
    \item an unrelated prompt entity must not be forced into $Y$;
    \item an unbindable or conflicting control should be ignored;
    \item the result must be photorealistic and physically coherent within the requested world.
\end{itemize}

The image model returns pixels only. Any internal interpretation of $X/Y$ is therefore unobserved; adherence is judged behaviorally from the output.

\paragraph{LLM-mediated semantic resolution}

Raw predicates such as \texttt{X\_held\_by\_Y} have role semantics inherited from their source contexts. A single global $X/Y$ assignment is often inappropriate when properties came from several topics.

For the Resolved variant, GPT-5.6 Sol receives:

\begin{itemize}
    \item the immutable primary prompt;
    \item the 20 sampled properties;
    \item instructions to retain as much sampled guidance as possible.
\end{itemize}

For each property it returns:

\begin{itemize}
    \item \texttt{use} or \texttt{drop};
    \item a local $X$ binding;
    \item a local $Y$ binding;
    \item a concrete visual instruction;
    \item a short rationale.
\end{itemize}

$X$ and $Y$ are local to one property. Different properties may bind them to different visible entities. The image model therefore receives concrete sentences rather than a globally inconsistent set of symbolic variables.

Sol returns one auditable JSON row per sampled property:

\begin{verbatim}
{
  "id": 0,
  "decision": "use",
  "x_binding": "the red ball",
  "y_binding": "an adult",
  "instruction": "An adult holds the red ball.",
  "reason": "compatible local binding"
}
\end{verbatim}

The concrete scene plan and retained instructions are then rendered under the same primary-prompt-priority contract as Direct.

For $\lambda=0.5$, Sol retained 9,920/10,000 controls (99.2\%). Thirty-two configurations required a documented moderation repair that removed minor-specific optional content and rebound safe properties to adults or objects.

\subsubsection{Experimental design}

\paragraph{Generation and evaluation settings.}
All three generators produce $1024\times1024$ images. GPT Image 2 uses its low-quality setting; FLUX.2-flex uses guidance 4.5 and 50 steps; MAI-Image-2.5 uses its default generation settings.
Independent and coordinated rewriting both preserve stated facts while varying composition, lighting, focus, viewpoint, and treatment. Repetitions are not filtered or replaced. The embedding models and RBF bandwidths are fixed across generators; bandwidths come from the original GPT Image 2 comparison. Odd-one-out uses one 40-image cascade per request and generator, with arm identities hidden and the pool reshuffled after each removal.

One MAI-Image-2.5 Direct output, candidate 7 for the parking-meter request ``A parking meter sitting in the middle of a flooded street.'', repeatedly triggered the provider's output-safety filter due to the sampled \texttt{Y\_child} attribute. For that image only, the renderer omitted the sampled \texttt{Y\_child} instruction. The original control remains in the sampling record, and the replacement is recorded in provenance. This case is outside the prespecified candidates 1 and 6 used for the validity and attribute audits.

\subsubsection{Exact evaluation prompts}

The experiment uses one frozen prompt from each of 50 distinct categories:

\begin{longtable}{r l l p{0.45\textwidth}}
    \toprule
    \textbf{\#} & \textbf{Category} & \textbf{Prompt ID} & \textbf{Exact prompt} \\
    \midrule
    \endhead
    1 & baseball bat & \texttt{baseball\_bat\_403675} & A guy holding a baseball bat is ready to play ball \\
    2 & cow & \texttt{cow\_537695} & A cow has a ring in its nose near two men. \\
    3 & wine glass & \texttt{wine\_glass\_285773} & A table with a wine glass and wine bottles on it \\
    4 & dining table & \texttt{dining\_table\_106338} & a big table that has some plates of food on it \\
    5 & spoon & \texttt{spoon\_306638} & A bowl of food set on a table with a spoon. \\
    6 & scissors & \texttt{scissors\_223362} & a close up of a person holding a pair of scissors \\
    7 & zebra & \texttt{zebra\_435569} & A giraffe and zebra are drinking at a pool of water. \\
    8 & vase & \texttt{vase\_157378} & a close up of flowers in a vase near a wall \\
    9 & bird & \texttt{bird\_548275} & A bird is perched on a rock in the sea water. \\
    10 & giraffe & \texttt{giraffe\_574402} & A baby giraffe is shown in the grass at the zoo. \\
    11 & train & \texttt{train\_448871} & A bullet train on an elevated track passing over parked cars. \\
    12 & potted plant & \texttt{potted\_plant\_546674} & A plant is growing in a clear vase filled with rocks \\
    13 & fork & \texttt{fork\_504711} & A closeup on some chocolate cake has a fork by it. \\
    14 & horse & \texttt{horse\_326058} & A bearded cowboy riding a horse in a dusty outdoor arena \\
    15 & suitcase & \texttt{suitcase\_137413} & a cat laying in a box next to an open suitcase \\
    16 & apple & \texttt{apple\_303937} & Apples and oranges sitting next to each other on a shelf. \\
    17 & stop sign & \texttt{stop\_sign\_158952} & A close-up of the stop sign and a street sign. \\
    18 & sink & \texttt{sink\_523426} & a bath room with a toilet a sink and a mirror \\
    19 & cup & \texttt{cup\_387434} & A cat drinking out of a cup on a night stand. \\
    20 & elephant & \texttt{elephant\_243418} & A city street with bicyclers and a man riding a elephant. \\
    21 & donut & \texttt{donut\_175604} & A donut seems to have lucky charms on top with coffee. \\
    22 & bear & \texttt{bear\_27272} & A bear is seen in the grass next to a creek. \\
    23 & sandwich & \texttt{sandwich\_20906} & A banana, peanut butter and turkey sandwich sitting in a container. \\
    24 & traffic light & \texttt{traffic\_light\_449895} & A construction worker installing and performing maintenance on a traffic light. \\
    25 & cake & \texttt{cake\_13605} & A 6 year old's birthday cake with horses on a top. \\
    26 & fire hydrant & \texttt{fire\_hydrant\_25809} & a close up of a fire hydrant on a side walk \\
    27 & backpack & \texttt{backpack\_137028} & a dog with shoes and a backpack standing next to car \\
    28 & frisbee & \texttt{frisbee\_218561} & A dog holds a frisbee with the side of its mouth \\
    29 & tennis racket & \texttt{tennis\_racket\_381108} & A man is hitting a tennis ball with a tennis racket. \\
    30 & truck & \texttt{truck\_123469} & A big truck going down the street with graffiti on it. \\
    31 & knife & \texttt{knife\_213881} & A carrot sitting on a cutting board next to a knife. \\
    32 & bench & \texttt{bench\_156466} & A bench is in front brick wall between two plant beds. \\
    33 & kite & \texttt{kite\_85911} & A kite has fallen from the air in to the water \\
    34 & sheep & \texttt{sheep\_273052} & A sheep lying on a hill side next to a tree. \\
    35 & bicycle & \texttt{bicycle\_174482} & A bicycle is chained to a fixture on a city street \\
    36 & teddy bear & \texttt{teddy\_bear\_546896} & A cat is curled up next to a light teddy bear. \\
    37 & boat & \texttt{boat\_135620} & A boat and people in the ocean with their sale boats. \\
    38 & parking meter & \texttt{parking\_meter\_234757} & A parking meter sitting in the middle of a flooded street. \\
    39 & baseball glove & \texttt{baseball\_glove\_468302} & a man is wearing a baseball glove at a game \\
    40 & handbag & \texttt{handbag\_70589} & a white handbag a camera a cigarette case and a flask \\
    41 & toilet & \texttt{toilet\_434098} & a bath room with a toilet a mirror and a sink \\
    42 & hot dog & \texttt{hot\_dog\_150026} & A close up view of a hot dog sandwich and beverage. \\
    43 & surfboard & \texttt{surfboard\_73946} & A couple of men standing on either side of a surfboard. \\
    44 & toothbrush & \texttt{toothbrush\_207060} & A magenta toothbrush sits next to a pair of pliers. \\
    45 & keyboard & \texttt{keyboard\_569272} & A cat laying down on a desk by a computer keyboard. \\
    46 & carrot & \texttt{carrot\_133050} & A knife sits next to a carrot on a cutting board. \\
    47 & oven & \texttt{oven\_203095} & a bunch of food is layed out next to an oven \\
    48 & hair drier & \texttt{hair\_drier\_416930} & A woman sitting under a hair dryer in a salon. \\
    49 & remote & \texttt{remote\_314188} & A cat laying on a bed next to a remote control. \\
    50 & motorcycle & \texttt{motorcycle\_410878} & a close up of a motorcycle parked on a dirt road \\
    \bottomrule
\end{longtable}

\subsubsection{Four-arm repeated odd-one-out removal}

Embedding metrics are corroborated by a blinded repeated odd-one-out evaluation over all 50 prompts.

For each prompt:

\begin{enumerate}
    \item pool ten prompt-only, ten independent-rewrite, ten coordinated-rewrite, and ten Topic-Ancestral MSR--Resolved images;
    \item deterministically shuffle the 40 images into one numbered montage;
    \item ask the vision judge to remove the single image most unlike the others;
    \item rebuild and reshuffle the montage;
    \item repeat for 30 rounds, leaving ten survivors.
\end{enumerate}

This uses 1,500 vision-judge rounds. Lower mean exit rank indicates greater diversity. Images surviving all 30 removals receive rank 40.
Results for all models are reported in Table~\ref{tab:t2i_vendi}.

\subsubsection{Validity and attribute realization.}

Separately, GPT-5.6 Sol judges prompt satisfaction, visual acceptability, and attribute realization for candidates 1 and 6 from each request and MSR variant: 100 images and 2,000 originally sampled attribute occurrences per variant and generator. An image is counted as valid only if it satisfies the request and is visually acceptable. 

Table~\ref{tab:t2i_audit} reports the recorded GPT-5.6 Sol judgments. Validity requires both request satisfaction and visual acceptability. Attribute-realization denominators include all 2,000 originally sampled occurrences in each audited arm, including controls omitted during resolution. The attribute judge receives the original attribute descriptions and generated image, but not the resolver's local role bindings. It can therefore interpret a role placeholder as the focal object rather than the supporting entity chosen by the resolver. These scores are model-based estimates of original-attribute realization, not a separate audit of fidelity to the resolved instructions.

\begin{table}[h]
\centering
\caption{Model-based audits of 100 images per MSR variant and generator. Each attribute-realization rate uses 2,000 originally sampled attribute occurrences.}
\label{tab:t2i_audit}
\begin{tabular}{lcccc}
\toprule
& \multicolumn{2}{c}{Valid images (\%)} & \multicolumn{2}{c}{Attributes realized (\%)} \\
\cmidrule(lr){2-3}\cmidrule(lr){4-5}
Generator & Direct & Resolved & Direct & Resolved \\
\midrule
GPT Image 2 & 99 & 97 & 61.6 & 80.2 \\
MAI-Image-2.5 & 96 & 97 & 65.3 & 78.5 \\
FLUX.2-flex & 95 & 90 & 71.1 & 77.9 \\
\bottomrule
\end{tabular}
\end{table}

The diversity gains coexist with generation errors. In the audited subset, 95--99\% of Direct images and 90--97\% of Resolved images meet both validity criteria (Table~\ref{tab:t2i_audit}). Resolution increases the recorded attribute-realization scores for every generator, but can reduce validity; the largest decrease is for FLUX.2-flex, from 95/100 to 90/100 valid images. To put these validity in numbers in perspective, note that in a deployment with reliable, inexpensive validity screening (an LLM validator is much cheaper than an image generator), unsuccessful outputs could be resampled. Assuming a constant per-attempt success probability of 90\%, this would require an expected $1/0.9\approx1.11$ generation attempts per valid output, corresponding to only about 11\% additional generation cost, plus the cost of screening.

\subsubsection{Lambda ablation}

For Direct rendering, increasing topic weight produces a monotonic diversity curve:

\begin{table}[h]
    \centering
    \begin{tabular}{rrr}
        \toprule
        $\lambda$ & \textbf{CLIP Vendi} & \textbf{DINOv2 Vendi} \\
        \midrule
        0.0 & 4.23 & 4.40 \\
        \textbf{0.5} & \textbf{5.14} & \textbf{4.97} \\
        1.0 & 5.34 & 5.21 \\
        \bottomrule
    \end{tabular}
\end{table}

With semantic resolution, the midpoint and topic endpoint are effectively tied under CLIP. The midpoint is higher under DINOv2:

\begin{table}[htbp]
    \centering
    \begin{tabular}{lrrrl}
        \toprule
        \textbf{Representation} & \textbf{$\lambda=.5$ Resolved} & \textbf{$\lambda=1$ Resolved} \\
        \midrule
        CLIP & 6.68 & 6.62 \\
        DINOv2 & 6.73 & 6.42 \\
        \bottomrule
    \end{tabular}
\end{table}

The midpoint therefore retains empirical statistical guidance without giving up the diversity of pure topic sampling.
\clearpage
\subsection{3D Folding Chairs}
\subsubsection{Data Collection}
\label{3d:data}
Collection began with a seven-image Amazon Berkeley Objects (ABO) \citep{collins2022abo} pilot. Its limited coverage motivated targeted searches for additional chair families and appearances: camping and director's chairs, deckchairs, transparent seating, shaped or painted timber, leather, cane, perforated panels, bamboo and historical scissor forms. Queries included named design examples such as Plia, Rex and Trieste, and non-English terms such as Klappstuhl, chaise pliante and sedia pieghevole.

The acquisition code scanned all 16 ABO metadata shards and queried Wikimedia Commons text search and chair categories, together with the Metropolitan Museum of Art, Art Institute of Chicago and Cleveland Museum APIs. The final corpus has 160 curated design/appearance records and 305 selected evidence images. It includes 159 folding-chair records and one explicitly identified demountable chair. The actual record counts by source are:

\begin{table}[h]
    \centering
    \renewcommand{\arraystretch}{1.2}
    \begin{tabular}{lr}
        \toprule
        \textbf{Source} & \textbf{Retained records} \\
        \midrule
        Amazon Berkeley Objects & 65 \\
        Wikimedia Commons & 83 \\
        Art Institute of Chicago & 1 \\
        Cleveland Museum of Art & 1 \\
        Metropolitan Museum of Art & 10 \\
        \bottomrule
    \end{tabular}
\end{table}
Only images with supported reusable licenses or public-domain declarations were admitted.

A model-assisted screening stage was followed by an investigator-assistant visual relevance pass over every retained core record's primary thumbnail, with targeted original-image checks. Clearly irrelevant scenes, fragments, backless stools, fixed auditorium seats and giant display replicas were excluded or held for review. Multiple views of a documented object were consolidated. The retained images were oriented, converted to RGB JPEG and limited to a 1,200-pixel longest edge; annotation inputs used at most 960 pixels.

\subsubsection{Mining recombinable properties and discovering correlations}
\paragraph{A part-specific matrix, including explicit unknowns}
Image-property extraction used the vision deployment gpt-5.6-sol. Astra performed the parent orchestration, implementation and source-relevance review; it was not the vision extractor. Discovery used a scaffold of frame, seat, back, arms, feet and hardware, with whole-chair relationships and accessories. Within each role, apparent material, color family, tone, finish, texture, construction and decorative pattern were separated. Additional visible structural and detail phrases were allowed. Thus the vocabulary was semi-structured and grounded in the images, rather than either wholly hand-enumerated or unconstrained free-form text.

Discovery phrases were consolidated into atomic definitions, with model-proposed mappings, validated coverage and explicit curator corrections. All retained records were then assessed against the same frozen dictionary in batches of three. Missing mentions were not treated as negatives. Backfill explicitly identified positives and uncertain properties, and assessed the remaining labels absent. After reconciling eight cross-axis aliases, the sampling-ready matrix contained 447 canonical properties, down from 455 assessed columns. Forty conflicting alias cells became unknown rather than being resolved arbitrarily.

\[
X[\text{item}, \text{property}] \in \{1: \text{observed present}, 0: \text{assessed absent}, \text{NA}: \text{uncertain or unseen}\}
\]

Cells with model-reported confidence below 0.75 remained unknown; this confidence is not calibrated probability. Unknown part presence gates that part's attributes to unknown, whereas known part absence makes its attributes false. Unreliable or grayscale color is unknown. About 85.5\% of cells are observed; 405 properties have at least one positive observation. Each of the 137 provisional family groups contributes total weight one, divided among its records, to reduce the influence of repeated retail variants and related views.

\paragraph{Correlations are design regularities, not mechanical laws}

The matrix exposes relationships at several scales: consistency of color and detailing across seat and back, associations between material appearance and member construction, and bundles of features characteristic of particular chair families. The following examples are drawn from the saved descriptive association table. They use only records where both properties are observed, with family weighting. A and B name positive properties; $\neg A$ means an assessed absence, not an unknown.

\begin{table}[htbp]
    \centering
    \renewcommand{\arraystretch}{1.2}
    \begin{tabular}{llrrr}
        \toprule
        \textbf{Property A} & \textbf{Property B} & \textbf{P(B $\mid$ A)} & \textbf{P(B $\mid$ not A)} & \textbf{Phi} \\
        \midrule
        Feet: metal & Frame: tubing & 95.3\% & 6.3\% & 0.890 \\
        Back: sling & Back: visible textile weave & 95.8\% & 8.5\% & 0.876 \\
        Back: light & Seat: light & 91.3\% & 3.8\% & 0.879 \\
        Back: edge binding & Seat: edge binding & 95.7\% & 11.8\% & 0.841 \\
        Back: geometric & Seat: geometric & 71.4\% & 2.2\% & 0.750 \\
        \bottomrule
    \end{tabular}
\end{table}

These associations are strong but deliberately non-perfect. The saved display filter required at least 15 known units of family weight and at least two in each joint/discordant positive cell, excluding same-axis coding pairs. For example, the relation between metal-looking feet and tubular frames reflects a recurrent material/construction vocabulary, while matched edge binding reflects a cross-part detailing convention. Neither relation says that alternative combinations are impossible. The gap between these regularities and hard constraints is precisely where recombination may be useful.

\paragraph{What the topic models revealed}

We first fit a six-component, family-weighted nonnegative factorization, minimizing reconstruction error only on observed entries. An initial diagnostic fit overemphasized sparsely observed packed-state properties, so the final eligibility rule required at least half the family weight to be observed, at least two weighted positives and negatives, and prevalence below 0.97. This retained 346 varying properties. Sparse properties remained in the full matrix; they were not recoded as absent.

A separate six-topic Bernoulli admixture model then provided an interpretable probabilistic inspector. Each chair has a topic mixture $\theta$; each topic has property probabilities $\beta$:

\[
P(X_{if} = 1 \mid \theta_i, \beta) = \sum_k \theta_{ik}\beta_{kf}
\]

Unknown cells are omitted from the likelihood, and records are family-weighted. The fit used Dirichlet/Beta priors and four seeded restarts. Six topics was an inspection choice, not a validated optimum. The labels below were assigned after fitting from enriched properties, not provided as training classes. Probabilities are fitted quantities, not held-out calibration results.

\begin{table}[htbp]
    \centering
    \renewcommand{\arraystretch}{1.3}
    \begin{tabular}{>{\raggedright\arraybackslash}p{0.35\textwidth} >{\raggedright\arraybackslash}p{0.6\textwidth}}
        \toprule
        \textbf{Post-hoc topic description} & \textbf{Two enriched properties: topic probability / corpus prevalence} \\
        \midrule
        Textile camping frames & Hardware: connector hubs: 95.2\% / 24.9\%; Arms: matte: 98.2\% / 35.8\% \\
        Timber slats and panels & Back: slats: 97.6\% / 17.3\%; Seat: wood: 96.8\% / 18.6\% \\
        Carved and weathered timber & Arms: brown: 97.5\% / 15.0\%; Feet: distressed: 93.0\% / 12.0\% \\
        Low striped slings & Continuous seat back: 96.6\% / 34.9\%; Low seat: 97.5\% / 36.3\% \\
        Laced lounge seating & Head cushion: 97.7\% / 13.0\%; Back: edge laced: 98.0\% / 13.5\% \\
        Solid-panel and molded seating & Back: solid panel: 94.7\% / 21.2\%; Seat: molded shell: 65.6\% / 8.5\% \\
        \bottomrule
    \end{tabular}
\end{table}

The qualitative story is more than a partition by color. The topics combine construction, attachment and surface treatment: textile camping frames with connector hubs and soft arm straps; timber slats; carved/weathered historical timber; low continuous striped slings; laced lounge chairs with head cushions; and smooth solid-panel/molded seating. Chairs can mix these topics. Such structure gives a statistical description of coherent design vocabularies while leaving room for new combinations.

\subsubsection{Model Output}
\label{3d:output}
The model was instructed with an intentionally short baseline design instruction:

\begin{tcolorbox}[
    enhanced,
    arc=1mm,                  %
    boxrule=1.5pt,             %
    colback=gray!5!white,      %
    colframe=gray!80!black,    %
    halign=center,             %
    width=\textwidth,       %
    center                     %
]
``Make a functional, interesting folding chair in 3D, including the hinges and other necessary mechanical components. Demonstrate it opening and closing. Explain what you actually checked and any limitations.'' %
\end{tcolorbox}

\begin{figure}[ht]
    \centering
    \includegraphics[width=\linewidth]{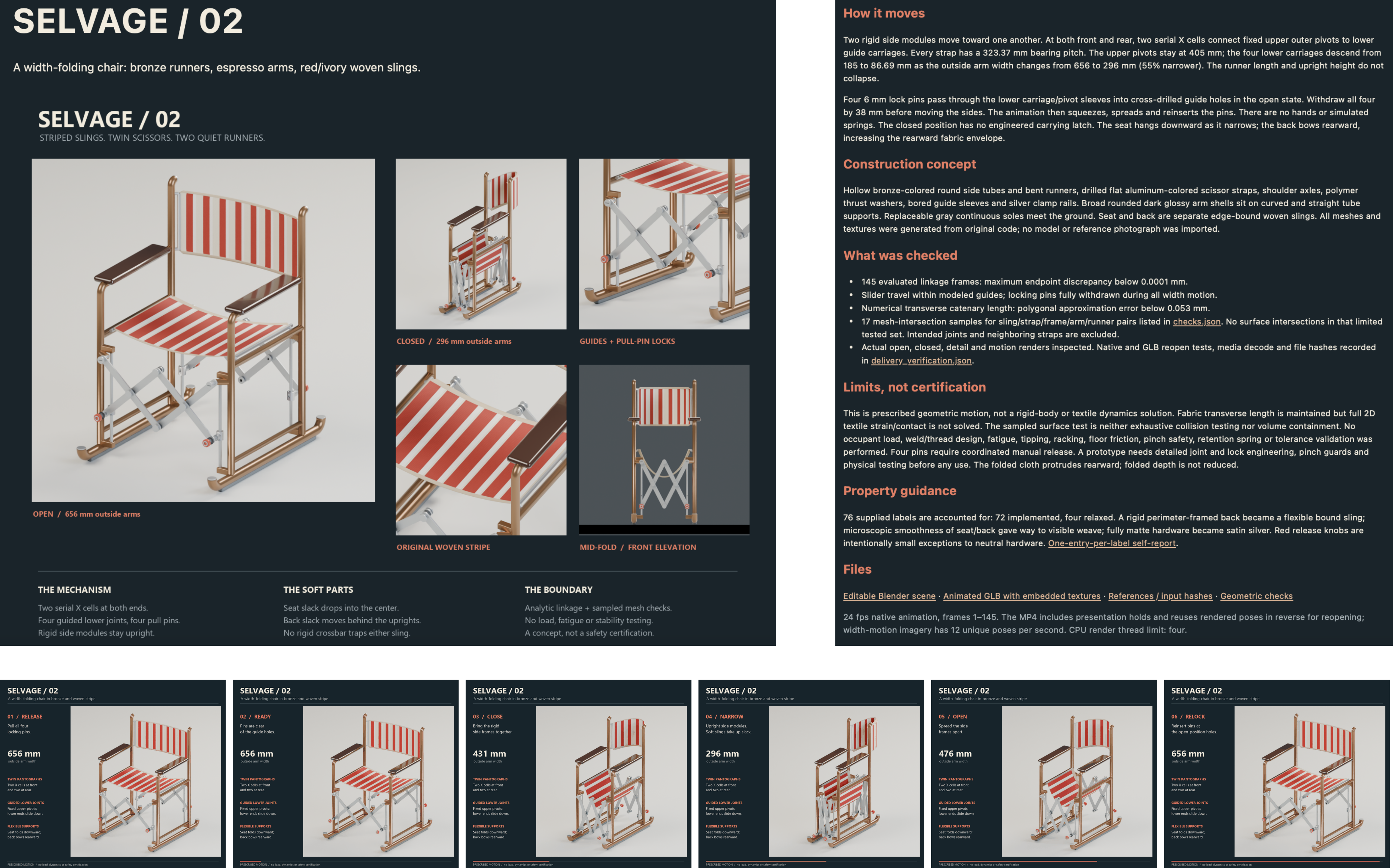}
    \caption{\textbf{(Top)} Example of a report generated by GPT-6 Astra for a generated folding chair. \textbf{(Bottom)} Illustration of the folding mechanism, originally provided as a video.}
    \label{fig:chair_report}
\end{figure}

The next paragraph required agents to construct geometry themselves rather than import chair or component models; to save reproducible source, an editable Blender scene, a GLB and demonstration outputs; and to record any references used. It explicitly left design, materials, construction method, and presentation to the agent. An example of a final report is provided in Figure~\ref{fig:chair_report}.

When using MSR, the attributes describe named chair parts---frame, seat, back, arms, feet, and hardware---and their materials, colors, textures, construction, and spatial relationships. Each sampled brief supplies 63--111 positive attributes; unlisted properties remain unspecified. For example, one configuration combined a blue textile sling seat and back, orange seat coloring, visible weave and edge binding, tubular crossed supports, paired ground rails, textile arm surfaces, and a cupholder. These controls suggest a design vocabulary, but do not specify dimensions, joint locations, or a folding mechanism.

\begin{figure}[ht]
    \centering
    \includegraphics[width=.8\linewidth]{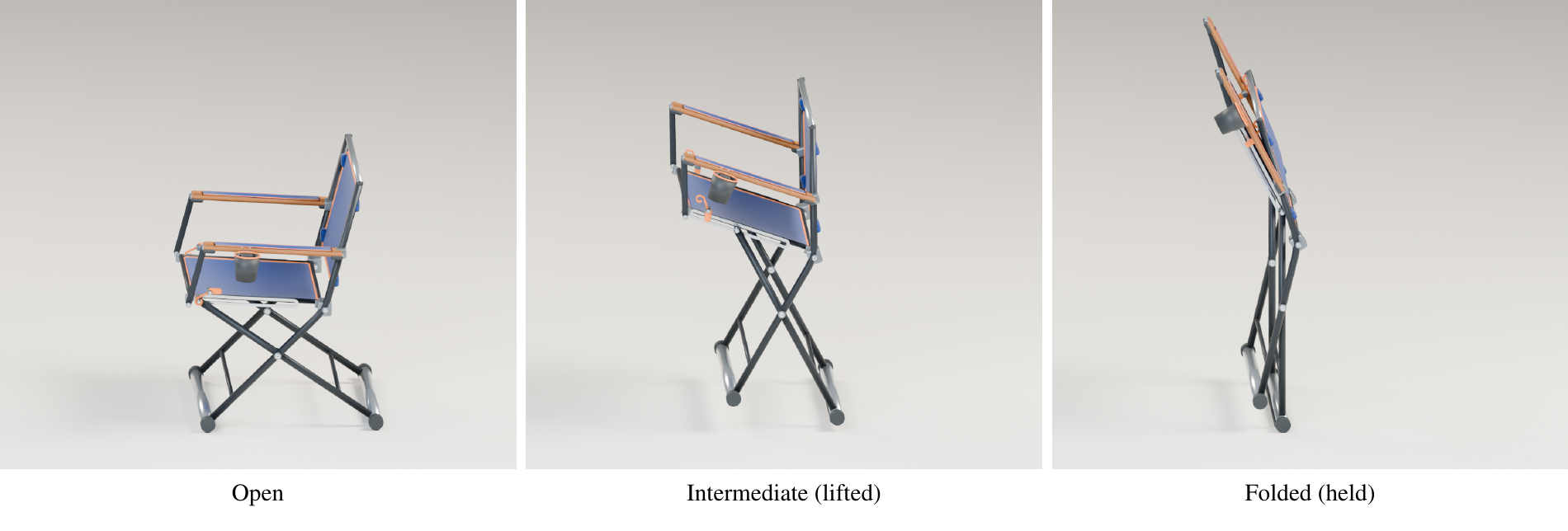}
    \caption{From sampled attributes to an articulated chair: open, intermediate, and folded states of the same generated Blender model. The programmed sequence includes a lift and tilt during folding; the folded pose is held rather than claimed to stand unsupported. These renders demonstrate modeled folding motion, not physically verified performance.}
    \label{fig:chair_folding}
\end{figure}

The agent must turn this guidance into an articulated 3D asset, not merely a plausible image. In this example, it authored a Blender scene with cobalt fabric and orange binding, scissor side frames, a hinged seat guided by sliding pins, and linked armrests (Figure~\ref{fig:chair_folding}). It also programmed a complete opening-and-closing sequence, including releasing the retaining hooks and lifting the chair during folding, and checked joint closure and selected component clearances across intermediate poses. Where the sampled properties conflicted with its realization, it replaced the proposed padded, open-mesh, continuous seat/back with separate unpadded textile slings and recorded those changes.

\subsubsection{Quality Evaluation}
\label{3d:eval}
 A fresh Astra judge received the common design assignment and neutral evaluation framing, but not the corpus, generation discussions, sampled labels, group identities, or expected result. Before viewing any candidate, it chose and froze five criteria, their weights, and scoring anchors. The investigator specified outcome-focused quality, forgiving concept validity, and the distinction between unusualness and quality; the judge chose the categories and weights.

The judge then reviewed anonymous original outcome images, detail views, full original demonstrations and motion sheets, with identity-stripped animated GLBs available for closer inspection. The actual quality evidence retained different cameras, lighting, and presentation styles. The judge was instructed to evaluate the chair rather than presentation polish or the generation process. It was not given source code, counts of checks, compliance claims, builder self-evaluations, or embedding scores.

Each criterion was scored by the same judge from 0 to 5 in half-point increments, with a weighted total out of 100. Table~\ref {tab:chair_quality} presents the per-criterion cohort means, recomputed from the saved chair-level scores.

\begin{table}[ht]
    \centering
    \small
    \begin{tabular}{l c c c}
        \toprule
        \textbf{Quality criterion} & \textbf{Weight} & \textbf{MSR} & \textbf{Coordinated}  \\
        \midrule
        Coherent deployed chair and support & 25\% & \textbf{4.40} & 3.70  \\
        Resolved, meaningful folding transformation & 30\% & \textbf{4.40} & 3.85 \\
        Human use, handling, and folded-state usefulness & 15\% & \textbf{3.90} & 3.35 \\
        Integrated form and detailed resolution & 20\% & \textbf{4.20} & 3.35 \\
        Compelling and purposeful design idea & 10\% & \textbf{3.90} & 3.75 \\
        \midrule
        \textbf{Weighted total / 100} & \textbf{100\%} & \textbf{84.70} & 72.55 \\
        \bottomrule
    \end{tabular}
    \caption{Evaluation of chair quality. Each criterion is scored out of 5 and assigned a weight by an investigator agent, different from the final judge.}
    \label{tab:chair_quality}
\end{table}
\clearpage
\section{Qualitative Examples}
\subsection{Webpage Generation}

\begin{figure}[h]
    \centering
    \includegraphics[width=.7\linewidth]{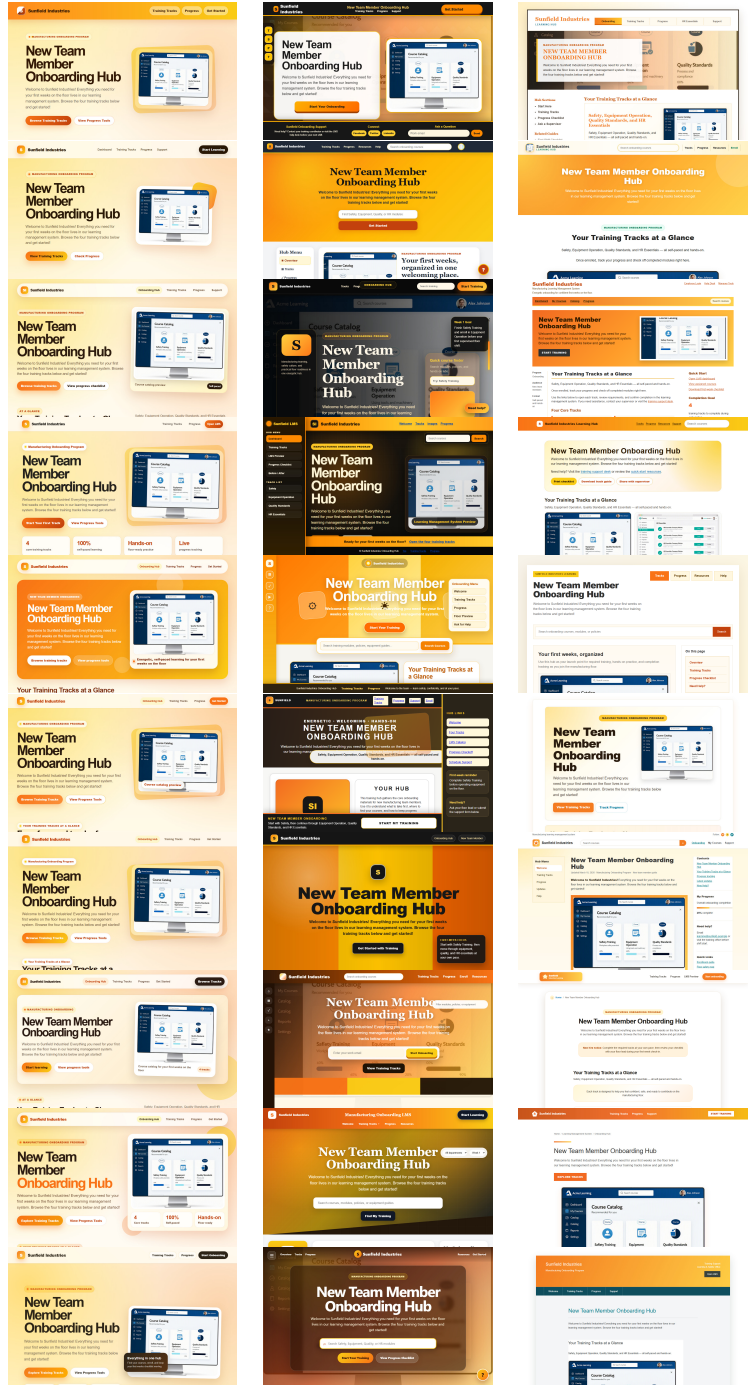}
    \caption{Webpages generated by GPT-5.5 using the prompt \textit{``Create a training hub page using the grounding file for the manufacturing onboarding program. Use orange and yellow as the main colors -- I want it to feel energetic and welcoming. Include all content and images from the grounding doc.''} \textbf{(Left)} Original prompt only. \textbf{(Middle)} \method{} using the Websight schema \textbf{(Right)} \method{} using the Design2Code schema.}
    \label{fig:webpage_gpt}
\end{figure}

\begin{figure}[h]
    \centering
    \includegraphics[width=.7\linewidth]{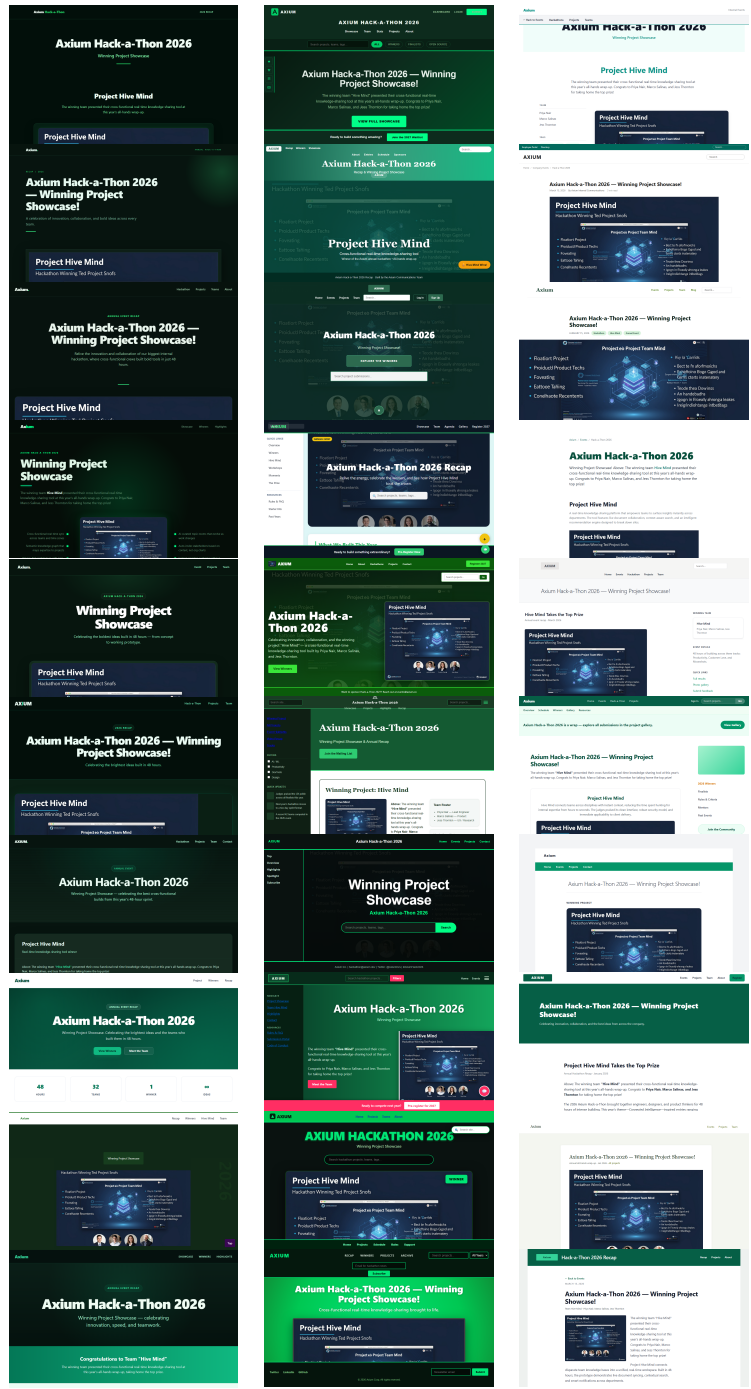}
    \caption{Webpages generated by Kimi K2.6 using the prompt \textit{``Create a recap page for the Axium annual hackathon using the grounding doc, green color scheme.''} \textbf{(Left)} Original prompt only. \textbf{(Middle)} \method{} using the Websight schema \textbf{(Right)} \method{} using the Design2Code schema.}
    \label{fig:webpage_kimi}
\end{figure}

\clearpage
\subsection{Text-to-Image}

\begin{figure}[h]
    \centering
    \includegraphics[width=\linewidth]{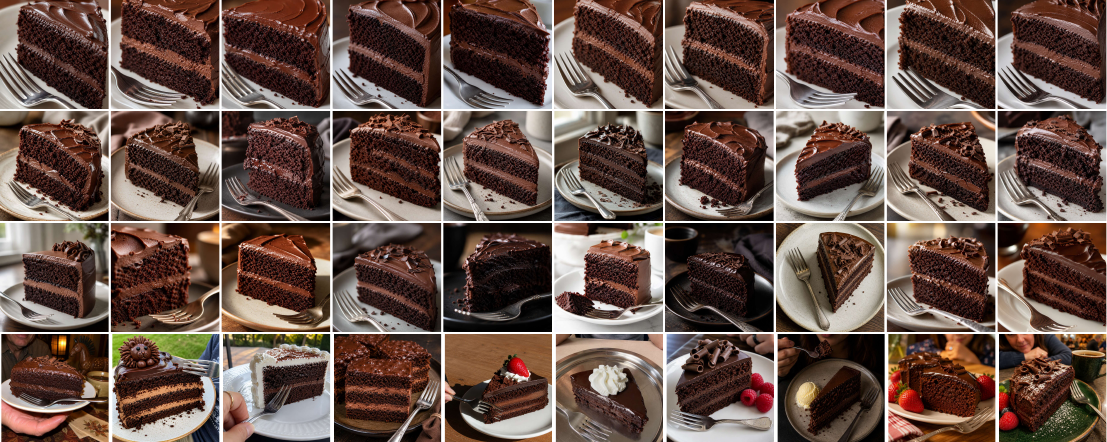}
    \caption{Images generated by GPT Image 2 using the prompt \textit{``A closeup on some chocolate cake has a fork by it.''} \textbf{(Row 1)} Original prompt only. \textbf{(Row 2)} Independent rewrite with GPT-5.6 Sol. \textbf{(Row 3)} Coordinated rewrite with GPT-5.6 Sol. \textbf{(Row 4)} \textsc{MSR}--Resolved.}
    \label{fig:t2i_cake}
\end{figure}

\begin{figure}[h]
    \centering
    \includegraphics[width=\linewidth]{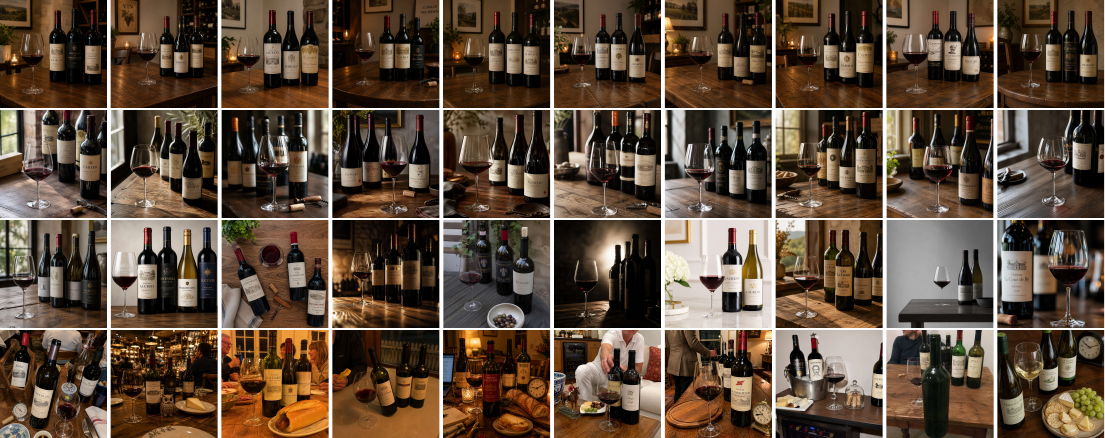}
    \caption{Images generated by GPT Image 2 using the prompt \textit{``A table with a wine glass and wine bottles on it.''} \textbf{(Row 1)} Original prompt only. \textbf{(Row 2)} Independent rewrite with GPT-5.6 Sol. \textbf{(Row 3)} Coordinated rewrite with GPT-5.6 Sol. \textbf{(Row 4)} \textsc{MSR}--Resolved.}
    \label{fig:t2i_wine}
\end{figure}

\begin{figure}[h]
    \centering
    \includegraphics[width=\linewidth]{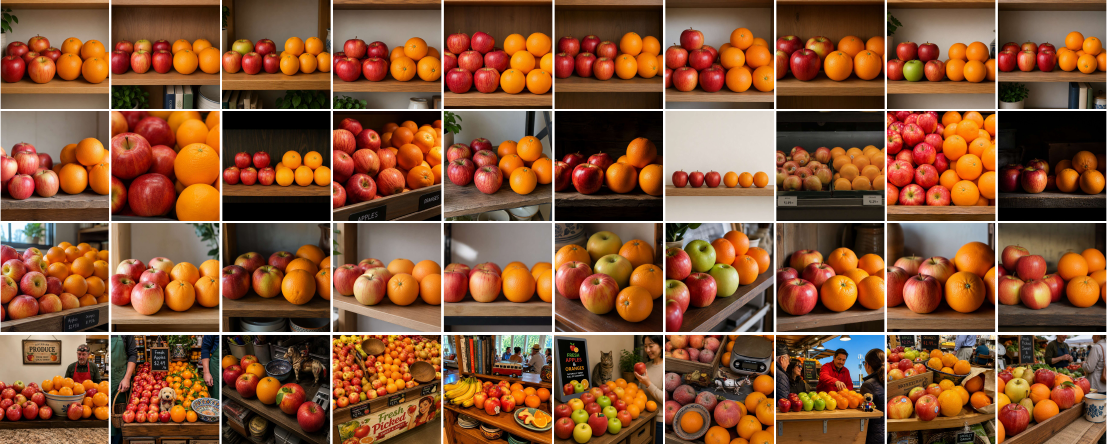}
    \caption{Images generated by GPT Image 2 using the prompt \textit{``Apples and oranges sitting next to each other on a shelf.''} \textbf{(Row 1)} Original prompt only. \textbf{(Row 2)} Coordinated rewrite with GPT-5.6 Sol. \textbf{(Row 3)} Independent rewrite with GPT-5.6 Sol. \textbf{(Row 4)} \textsc{MSR}--Resolved.}
    \label{fig:t2i_apple}
\end{figure}
\clearpage
\subsubsection{Out-of-Distribution examples}
\begin{figure}[ht]
    \centering
    \includegraphics[width=\linewidth]{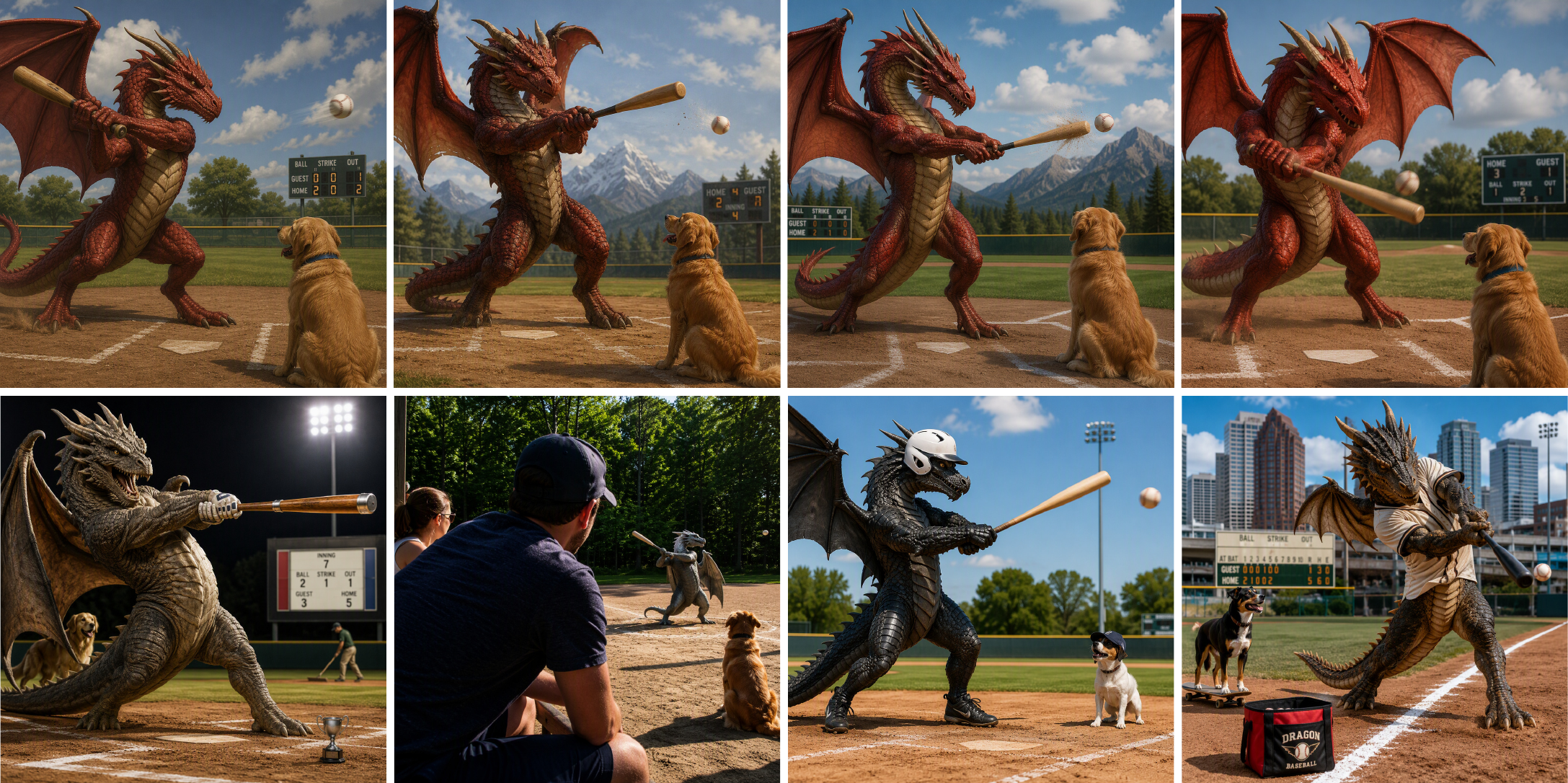}
    \caption{Images generated by GPT Image 2 using the prompt \textit{``A dragon swinging a bat with a dog watching.''} \textbf{(Row 1)} Original prompt. \textbf{(Row 2)} \method{}.}
    \label{fig:t2i_ood_dragon}
\end{figure}
Even though the schema was mined from images from the COCO dataset, many mined labels capture information that is more general: about composition, layering, camera angles, and certain subject co-occurrence. Therefore, our schema is useful even for generating images outside of the distribution we aligned to - the one hundred COCO categories (Figure \ref{fig:t2i_ood_dragon}). Also note that the generative model, given the prompt and the attributes, does its best to honor both.

\end{document}